\documentclass[lettersize,journal]{IEEEtran}
\usepackage{amsmath,amssymb,amsfonts}
\usepackage{algorithmic}
\usepackage{algorithm}
\usepackage{array}
\usepackage[caption=false,font=normalsize,labelfont=sf,textfont=sf]{subfig}
\usepackage{textcomp}
\usepackage{stfloats}
\usepackage{url}
\usepackage{verbatim}
\usepackage{graphicx}
\usepackage{cite}

\usepackage{algorithmic}
\usepackage{xcolor}
\usepackage{multirow}
\usepackage{threeparttable}
\usepackage{array}
\usepackage{booktabs}
\usepackage{amsbsy}
\begin{document}

\title{LiDAR-Inertial Odometry in Dynamic Driving Scenarios using Label Consistency Detection}

\author{Zikang~Yuan$^{1}$, Xiaoxiang~Wang$^{2}$, Jingying~Wu$^{2}$, Junda~Cheng$^{2}$ and Xin~Yang$^{2*}$% <-this % stops a space
	\thanks{$^{1}$Zikang~Yuan is with Institute of Artificial Intelligence, Huazhong University of Science and Technology, Wuhan, 430074, China. (E-mail: {\tt\small yzk2020@hust.edu.cn})}%
	\thanks{$^{2}$Xiaoxiang~Wang, Jingying~Wu, Junda~Cheng and Xin~Yang$^{*}$ are with the Electronic Information and Communications, Huazhong University of Science and Technology, Wuhan, 430074, China. (* represents the corresponding author. E-mail: {\tt\small m202272556@hust.edu.cn; m202177065@hust.edu.cn; jundacheng@hust.edu.cn; xinyang2014@hust.edu.cn})}%
}

\markboth{Journal of \LaTeX\ Class Files,~Vol.~14, No.~8, August~2021}%
{Shell \MakeLowercase{\textit{et al.}}: A Sample Article Using IEEEtran.cls for IEEE Journals}


\maketitle

\begin{abstract}
In this paper, a LiDAR-inertial odometry (LIO) method that eliminates the influence of moving objects in dynamic driving scenarios is proposed. This method constructs binarized labels for 3D points of current sweep, and utilizes the label difference between each point and its surrounding points in map to identify moving objects. Firstly, the binarized labels, i.e., ground and non-ground are assigned to each 3D point in current sweep using ground segmentation. In actual driving scenarios, dynamic objects are always located on the ground. For most points scanned from moving objects, they cannot coincide with any existing structures in space. For a minority of moving objects' points that are close to the ground, their labels exhibit differences with surrounding ground points. Thus, the points on moving objects are identified due to lacking of nearest neighbors in map or inconsistency with the labels of surround ground points. The nearest neighbors from global map are localized by voxel-location-based nearest neighbor search and the consistency is evaluated by comparing the label consistency with nearest neighbors, without involving any massive computations. Finally, the points on moving objects are removed. The proposed method is embeded into a self-developed LIO system (i.e., Dynamic-LIO), evaluated with six public datasets, and tested in both dynamic and static environments. Experimental results demonstrate that our method can identify moving objects with extremlely low computational overhead (i.e., 1$\sim$9ms/sweep), and our Dynamic-LIO can achieve state-of-the-art pose estimation accuracy in both static and dynamic scenarios. We have released the source code of this work for the development of the community.
\end{abstract}


\section{Introduction}
\label{Introduction}

\begin{figure}
	\begin{center}
		\includegraphics[scale=0.72]{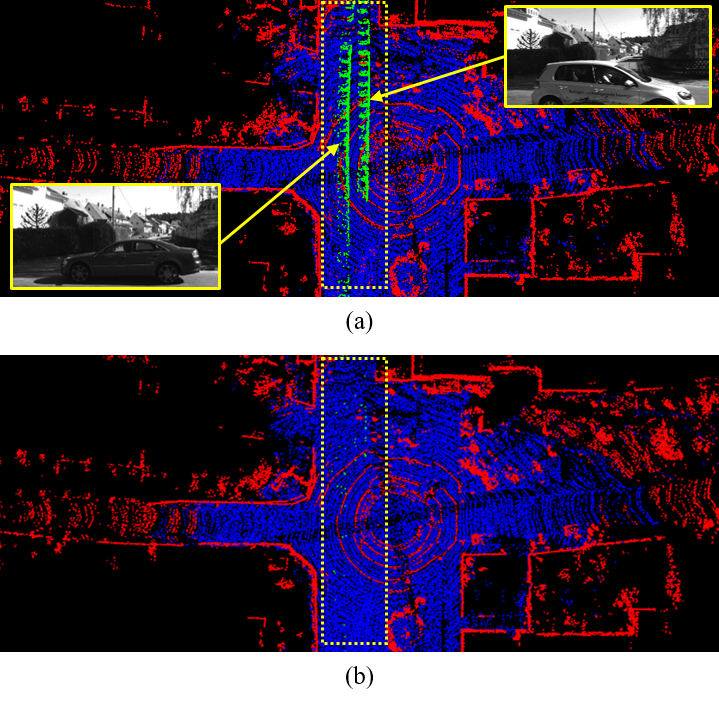}
		\caption{Illustration of (a) the exemplar point cloud map with dynamic points, where green points are ghost tracks of moving vehicles. (b) the static point cloud map, where the dynamic points have been detected and removed by our label consistency detection method.}
		\label{fig1}
	\end{center}
\end{figure}

\IEEEPARstart{I}{n} recent years, 3D light detection and ranging (LiDAR) based state estimation methods, including LiDAR-inertial odometry (LIO), have played an important role in autonomous driving thanks to the strong strength of LiDAR to perceive 3D space. Theses methods can provide 15 degree-of-freedom (DOF) state estimation of vehicle platform and recover 3D structure of surrounding environment in real time. However, the scenarios where they apply are strictly limited by the assumption of static environment. In actual driving scenarios, moving vehicles or pedestrians can leave ghost tracks in the map (as shown in Fig. \ref{fig1} (a)), leading to cumulative errors in state estimation and providing erroneous observational information for obstacle avoidance. Given the necessity for real-time state estimation and mapping in a LIO system, the computational cost of removing dynamic objects must be within the processing time budget for a single sweep. Hence, efficiently identifying moving objects from 3D point clouds is crucial.

To address the issues of poor map reconstruction and inaccurate state estimation caused by ghost tracks, researchers employ a range of approaches such as point correlation, visibility, occupancy probability and semantic information to identify and remove dynamic 3D points from input LiDAR sweeps. For point correlation \cite{dai2020rgb}, the correlation exists between static points, but there is no correlation between dynamic and static points. The connectivity of map points can be exploited to separate moving objects from the static environments. However, the construction and maintenance of a correlation graph involve computing pairwise Euclidean distances of batch 3D points, and the entire process requires substantial computational resources. For visibility \cite{kim2020remove, yoon2019mapless, fan2022dynamicfilter}, the static pixels on re-projected image planes remain invariant across multiple sweeps, whereas those dynamic pixels undergo displacements. However, the generation of re-projected image plane entails a computational process to map large number of LiDAR points from their original 3D space onto the 2D image plane. For occupancy probability \cite{schmid2023dynablox, yan2023rh}, the occupancy status of voxels corresponding to static environments remains constant over time, whereas the occupancy status of voxels occupied by dynamic objects will change with time. However, the estimation of an occupancy grid map need to combine multiple nearest static submaps and count the occupancy status of each voxel in each submap. The three aforementioned methods leverage extensive quantitative geometric computations and global statistics to identify moving objects in 3D point clouds, and they incur high computational costs, which limits their integration into LIO systems. Deep learning methods \cite{cortinhal2020salsanext, milioto2019rangenet} use learned semantic information for rapid qualitative separation of dynamic objects. However, they depend on extensive labeled data, risk failure with unlabeled classes, and necessitate powerful Graphic Processing Units (GPUs) for real-time operation, which can detract from resources available for subsequent tasks such as trajectory planning.

In contrast to existing approaches involving extensive quantitative geometric computations, global statistics or prior semantic information to classify moving objects and static environments, in this study, a qualitative identification criterion based on label difference with nearest neighbors is used to identify 3D points on moving objects. Since moving vehicles and pedestrians are both situated on the ground, the potential dynamic points only exist in non-ground space, while the bases of moving objects are adjacent to the ground. Based on this characteristic, a label consistency detection method, which can fastly identify moving objects without any prior semantic informations, is proposed to classify moving objects and static environments in driving scenarios. The core idea of the proposed label consistency detection method consists of two parts. First, a fast 2D connected components \cite{himmelsbach2010fast} is utilized to divide the 3D points of current sweep into binarized labels, i.e., ground points and non-ground points. Then, dynamic points are determined by comparing label consistency with nearest neighbors, which are directly localized by a voxel-location-based nearest neighbor search method. The whole dynamic point identification process including voxel-location-based nearest neighbor search and label consistency comparison, is devoid of any operations about geometric computations or global statistics. This characteristic ensures that the proposed method has the advantage of a expremely low computational overhead (i.e., 1$\sim$9ms/sweep). Finally, a self-developed LIO system (i.e., Dynamic-LIO) that uses this label consistency detection method is proposed to eliminate the influence of moving objects in driving scenarios. In order to guarantee that the final reconstructed map comprises solely static points, any detected dynamic points will be excluded from the global map (as shown in Fig. \ref{fig1} (b)). Subsequently, the clean static global map can be used to perform label consistency detection for the next sweep, ensuring the sustainability of the proposed method.

Experimental results on three public datasets of dynamic environments, i.e., $semantic$-$kitti$ \cite{behley2019semantickitti}, $ulhk$-$CA$ \cite{wen2020urbanloco} and $urban$-$Nav$, demonstrate that our method achieves comparable preservation rate (PR), rejection rate (RR) and absolute trajectory error (ATE), and sighnificantly outperforms online dynamic point detection methods \cite{fan2022dynamicfilter, yan2023rh, qian2022rf, wu2023lidar} in terms of expremely low computational overhead (1$\sim$9ms/sweep). In addition, experimental results on three public datasets of static environments, i.e., $nclt$ \cite{carlevaris2016university}, $utbm$ \cite{yan2020eu} and $ulhk$-$HK$ \cite{wen2020urbanloco}, demonstrate that our Dynamic-LIO also outperforms state-of-the-art LIO systems for static scenes in terms of smaller ATE.

To summarize, the main contributions of this work are three aspects: 1) We propose a label consistency detection method for fast identification of 3D dynamic points. It circumvents the operations of extensive geometric computations or global statistics and thus achieves lightweight; 2) We develop a LIO system and integrate the proposed label consistency detection method into this LIO in a unified manner, improving the accuracy of pose estimation in dynamic scenes by eliminating the ghost tracks in the reconstructed map; 3) We have released the source code of our approach to facilitate the development of the community\footnote{https://github.com/ZikangYuan/dynamic\_lio}.

The rest of this paper is structured as follows. Sec. \ref{Related Work} reviews the relevant literature. Sec. \ref{Preliminary} provides preliminaries. Secs. \ref{Label Consistency Detection} details our label consistency detection method and Sec. \ref{Our System Dynamic-LIO} introduces our system Dynamic-LIO, followed by experimental evaluation in Sec. \ref{Experiments}. Sec. \ref{Conclusion} concludes the paper.

\section{Related Work}
\label{Related Work}

In this section, we review the related work about existing 3D dynamic point detection approaches without deep learning \cite{pomerleau2014long, dai2020rgb, yoon2019mapless, kim2020remove, hornung2013octomap, schauer2018peopleremover, lim2021erasor, chen2023dorf, schmid2023dynablox, yan2023rh} and current mainstream LIO systems for static and dynamic scenes \cite{shan2020lio, qin2020lins, xu2021fast, he2023point, chen2023direct, chen2024ig, yuan2023semi, yuan2022sr, pfreundschuh2021dynamic, qian2022rf, wu2023lidar, xu2024lidar, zhang2024as}. Although there are some deep learning based 3D dynamic point detection methods \cite{cortinhal2020salsanext, milioto2019rangenet}, they are weakly relevant to this work so we omit the detailed discussion of them.

Pomerleau el. al. \cite{pomerleau2014long} utilized the motion pattern of point to represent the correlation. They calculated the motion pattern of each 3D point and infer the dominant motion patterns within the map, then determined the points that do not fit the motion pattern as dynamic points. Dai et. el. \cite{dai2020rgb} utlized the relative position between two points to represent the correlation, and then utilized the amplitude of relative position change over time as the criterion of consistency to identify dynamic points. However, the computational overhead of calculating motion pattern and maintaining map point correlation in large-scale outdoor scenarios is prohibitive. Yoon et. al. \cite{yoon2019mapless} proposed to simply query one sweep against another, and identify point with evident visibility difference as dynamic point. Removert \cite{kim2020remove} proposed a multi-resolution range image-based false prediction reverting algorithm. This method first conservatively retained definite static points and iteratively recover more uncertain static points by enlarging the query-to-map association window size. However, visibility-based approaches usually suffers from incidence angle ambiguity and occlusion issues. In addition, the generation of multiple projections on the spherical image plane and the assignment of a static value to each point in visibility-based approaches require high computational overhead. OctoMap \cite{hornung2013octomap} firstly proposed a framework to generate volumetric 3D environment model, which is based on octrees and uses probabilistic occupancy estimation. Given a registered set of 3D points, Schauer et. al. \cite{schauer2018peopleremover} build a regular voxel occupancy grid and then traverse it along the lines of sight between the sensor and the measured points to identify the differences in volumetric occupancy between multiple sweeps. Erasor \cite{lim2021erasor} proposed the concept called pseudo occupancy to express the occupancy of unit space and then discriminate spaces of varying occupancy. Then, the region-wise ground plane fitting (R-GPF) method is adopted to distinguish static points from dynamic points within the candidate bins that potentially contain dynamic points. DORF \cite{chen2023dorf} proposed a novel coarse-to-fine offline framework that exploits global 4D spatial-temporal LiDAR information to achieve clean static point cloud map generation. DORF first conservatively preserved the definite static points leveraging the receding horizon sampling (RHS) mechanism, then gradually recovered more ambiguous static points, guided by the inherent characteristic of dynamic objects in urban environments. \cite{schmid2023dynablox} proposed to incrementally estimate high confidence free-space areas by modeling and accounting for sensing, state estimation, and mapping limitations during online robot operation. It can achieve robust moving object detection in complex unstructured environments. RH-Map \cite{yan2023rh} proposed a novel map construction framework based on 3D region-wise hash map structure, which adopts the two-layer 3D region-wise hash map structure and the region-wise ground plane estimation for dynamic object removal. Occupancy map-based approaches are usually accompanied by the nearest neighbor search, confidence calculation, occupancy probability statistics, relative spatial position calculation and other operations requiring batch geometric computation, which cause a significant computational burden.

In recent years, various LIO systems have been proposed in the robotics community. LIO-SAM \cite{shan2020lio} firstly formulated LIO as a factor graph, which allows the incorporation of a multitude of relative and absolute measurements, including loop closures, as factors from different sources into the system. In LINs \cite{qin2020lins}, a pioneering integration of 6-axis IMU and 3D LiDAR was accomplished within an error state iterated Kalman filter (ESIKF) framework. This design ensures that the computational demands of the system remain tractable. Based on mathematical foundations, Fast-LIO \cite{xu2021fast} adapted a technique of solving Kalman gain \cite{sorenson1966kalman}, circumventing the need for high-order matrix inversion, thereby significantly alleviating the computational load. Building upon the advancements of Fast-LIO, Fast-LIO2 \cite{xu2022fast} introduced an innovative ikd-tree algorithm \cite{cai2021ikd}. Compared to the conventional kd-tree, this algorithm offers reduced temporal expenditure in processes such as tree construction, traversal, and element removal. Point-LIO \cite{he2023point} proposed a point-by-point LIO framework that updates the state at each LiDAR point measurement, which allows an extremely high-frequency output. DLIO \cite{chen2023direct} proposed to preserve a third-order minimum within the realms of state prediction and point distortion calibration, thereby facilitating the acquisition of more precise pose estimation. IG-LIO \cite{chen2024ig} integrated the generalized-ICP (GICP) constraints and inertial constraints into a unified estimation framework. In addition, iG-LIO employed a voxel-based surface covariance estimator to estimate the surface covariances of scans, and utilized an incremental voxel map to represent the probabilistic models of surrounding environments. Semi-Elastic-LIO \cite{yuan2023semi} proposed a semi-elastic optimization-based LiDAR-inertial state estimation method, which imparts sufficient elasticity to the state to allow it be optimized to the correct value. SR-LIO \cite{yuan2022sr} adapted the sweep reconstruction method \cite{yuan2023sdv, yuan2024sr}, which segments and reconstructs raw input sweeps from spinning LiDAR to obtain reconstructed sweeps with higher frequency. Consequently, the frequency of estimated pose is also increased. Pfreundschuh et. al. \cite{pfreundschuh2021dynamic} proposed an end-to-end occupancy grid based pipeline that can automatically label a wide variety of arbitrary dynamic objects, and embeded this network into a LiDAR odometry system. RF-LIO \cite{qian2022rf} utilized an adaptive multi-resolution range images to first remove dynamic objects, and then match LiDAR sweeps to the map for state estimation. ID-LIO \cite{wu2023lidar} proposed a LiDAR-inertial odometry based on indexed point and delayed removal strategy for dynamic scenes, which builds on LIO-SAM. Although RF-LIO and ID-LIO have the ability to perform state estimation in dynamic scenarios, huge computational overhead makes them unable to run stably in real time.

\section{Preliminary}
\label{Preliminary}

\subsection{Coordinate Systems}
\label{Coordinate Systems}

We denote $(\cdot)^w$, $(\cdot)^l$ and $(\cdot)^b$ as a 3D point in the world coordinate, the LiDAR coordinate and the IMU coordinate respectively. The world coordinate is coinciding with $(\cdot)^b$ at the starting position.

We denote the LiDAR coordinate for taking the $i_{th}$ sweep at time $t_i$ as $l_i$ and the corresponding IMU coordinate at $t_i$ as $b_i$, then the transformation matrix (i.e., external parameters) from $l_i$ to $b_i$ is denoted as $\mathbf{T}_{l_i}^{b_i} \in S E(3)$, which consists of a rotation matrix $\mathbf{R}_{l_i}^{b_i} \in S O(3)$ and a translation vector $\mathbf{t}_{l_i}^{b_i} \in \mathbb{R}^3$. The external parameters are usually calibrated once offline and remain constant during online pose estimation. Therefore, we can represent $\mathbf{T}_{l_i}^{b_i}$ as $\mathbf{T}_{l}^{b}$ for simplicity. The pose from the IMU coordinate $(\cdot)^{b_i}$ to the world coordinate $(\cdot)^{w}$ is strictly defined as $\mathbf{T}_{b_i}^{w}$.

\begin{figure*}
	\begin{center}
		\includegraphics[scale=0.31]{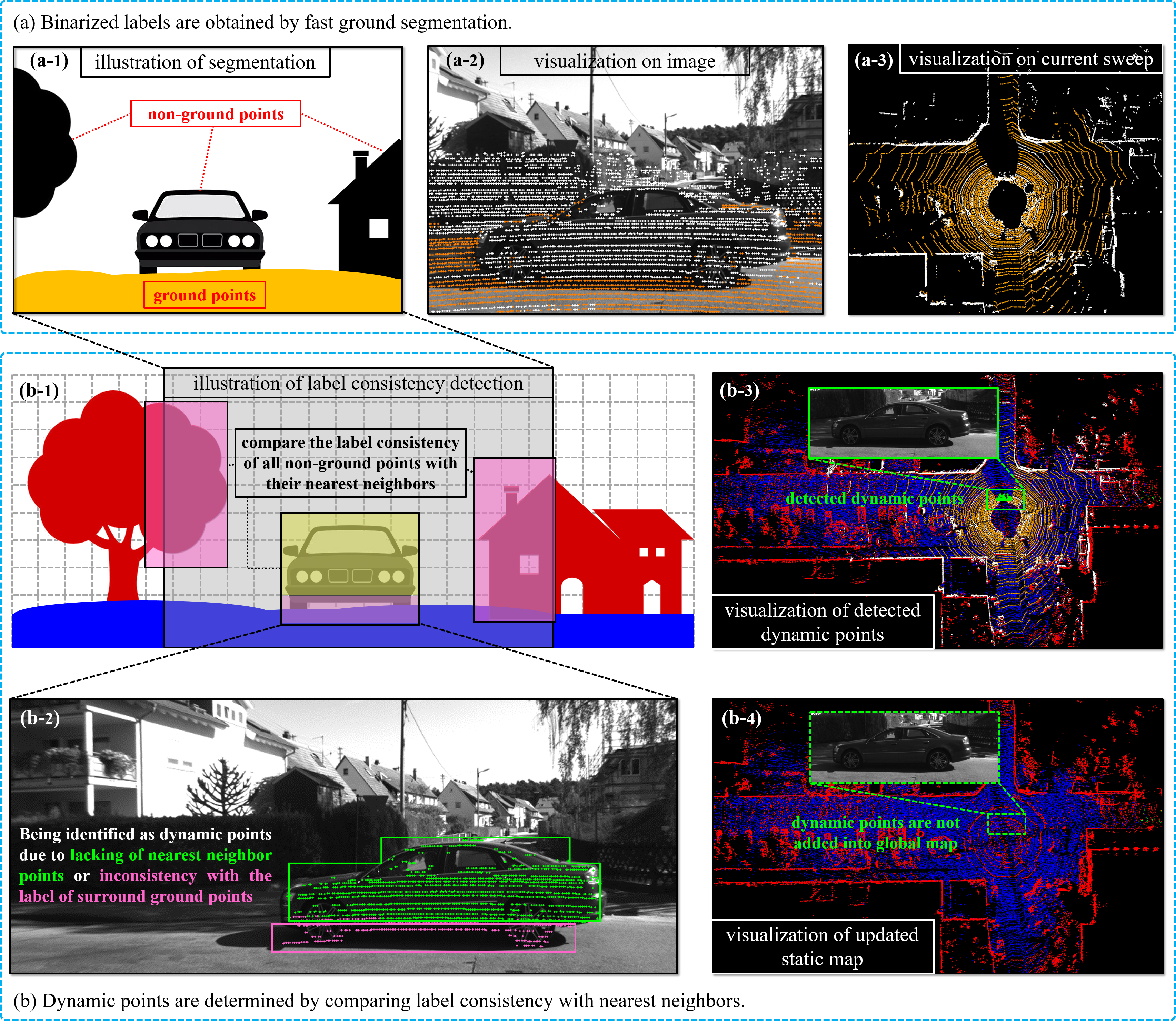}
		\caption{Illustration and examplar results of the proposed label consistency detection method. (a) Binarized labels are obtained by fast ground segmentation, where all points are divided into ground points (i.e., orange points in (a-2) and (a-3)) and non-ground points (i.e., white points in (a-2) and (a-3)). (b) Dynamic points are determined by comparing label consistency with nearest neighbors. The grid space where the dynamic point is located at the current time was not occupied in the past (i.e., green area in (b-1)). Thus the green points in (b-2) are detected as dynamic points because they are unable to find enough nearest neighbors. The occupied grid space does not exclusively contain static points (i.e., pink area in (b-1)), there may be some dynamic points adjacent to the ground. Thus the pink points in (b-2) are detected as dynamic points because their labels are inconsistent with surround ground points. The blue points and red points in (b-3) and (b-4) are ground and non-ground points in global map respectively.}
		\label{fig4}
	\end{center}
\end{figure*}

\subsection{Voxel map Management}
\label{Voxel map Management}

The entire system maintains two global maps: the tracking-map and the output map. The former is utilized for state estimation, while the latter is utilized for label consistency detection and serves as the final reconstruction outcome. In Dynamic-LIO, the tracking-map has already removed out the vast majority of dynamic points. However, to prevent over-filtering that could lead to insufficient geometric information for LIO, we refrain from further processing the tracking-map and instead focus on the output map (as illustrated in Sec. \ref{Undetermined-Point Re-Determination}). Thus compared to the tracking-map, the dynamic points in the output map are removed more thoroughly. Both the tracking-map and the output map are managed by voxel, whose voxel resolution is $1.0\times1.0\times1.0$ (unit: m) and each voxel contains a maximum of 20 points.

\section{Label Consistency Detection}
\label{Label Consistency Detection}

Label consistency detection aims to circumvent batch geometric computations and global statistics, which are prevalent in existing dynamic point detection approaches, thereby facilitating rapid identification of 3D dynamic points. To our knowledge, most existing methods primarily involve batch geometric computations and global statistics in aspects of nearest neighbor search and consistency evaluation. Consequently, we are committed to achieve the lightweight of these two aspects in our method.

The core premise of label consistency detection is that the moving objects in driving scenarios are in contact with the ground. Under this premise, we first construct the binarized labels (i.e., ground label and non-ground label) for each 3D point by segmenting the ground points from current input sweep (as illustrated in Fig. \ref{fig4} (a)). All ground points are inherently static, and the potential dynamic points are exclusively found among non-ground points. If we have already prepared the static global map at the previous moment, aside from the new points to be added at a greater distance, each static point at current moment can find its corresponding nearest neighbor within the global map during map update. For LiDAR points scanned from moving objects, the lack of structural informations within the global map prevents the current position from coinciding with any existing static geometric structures in space. (as illustrated in the green area of Fig. \ref{fig4} (b-1)). Thus, most LiDAR points scanned from moving objects are often unable to find nearest neighbors during registration and we identify thoes points as dynamic points (shown as the green points in Fig. \ref{fig4} (b-2)). As for the remaining small subset of LiDAR points (shown as the pink points in Fig. \ref{fig4} (b-2)), they may find ground points as their nearest neighbors. We then determine whether to classify them as dynamic points according to the proportion of ground points within the nearest neighbors. It is evident that throughout the process of evaluating label consistency, we only need to calculate the proportion of ground points among the nearest neighbors, without engaging in any batch geometric computations and global statistics. In addition, we utilize the voxel-location-based nearest neighbor search to obtain nearest neighbors, which can be directly located without any quantitative geometric distance calculation. The lightweight of these two core aspects ensures the low computational overhead of our method. Once the dynamic points at the current moment are identified, we utilize the estimated pose from LIO to register the static points into the global map to finish the map update, which can be used to identify dynamic points of next sweep, thereby ensuring the sustainability of our method.

Specially, the label consistency detection method is divided into five steps: binarized label construction, background separation, voxel-location-based nearest neighbor search, dynamic point determination and undetermined-point re-determination. In the following, we will provide a detailed description of each step.

\begin{figure}
	\begin{center}
		\includegraphics[scale=0.445]{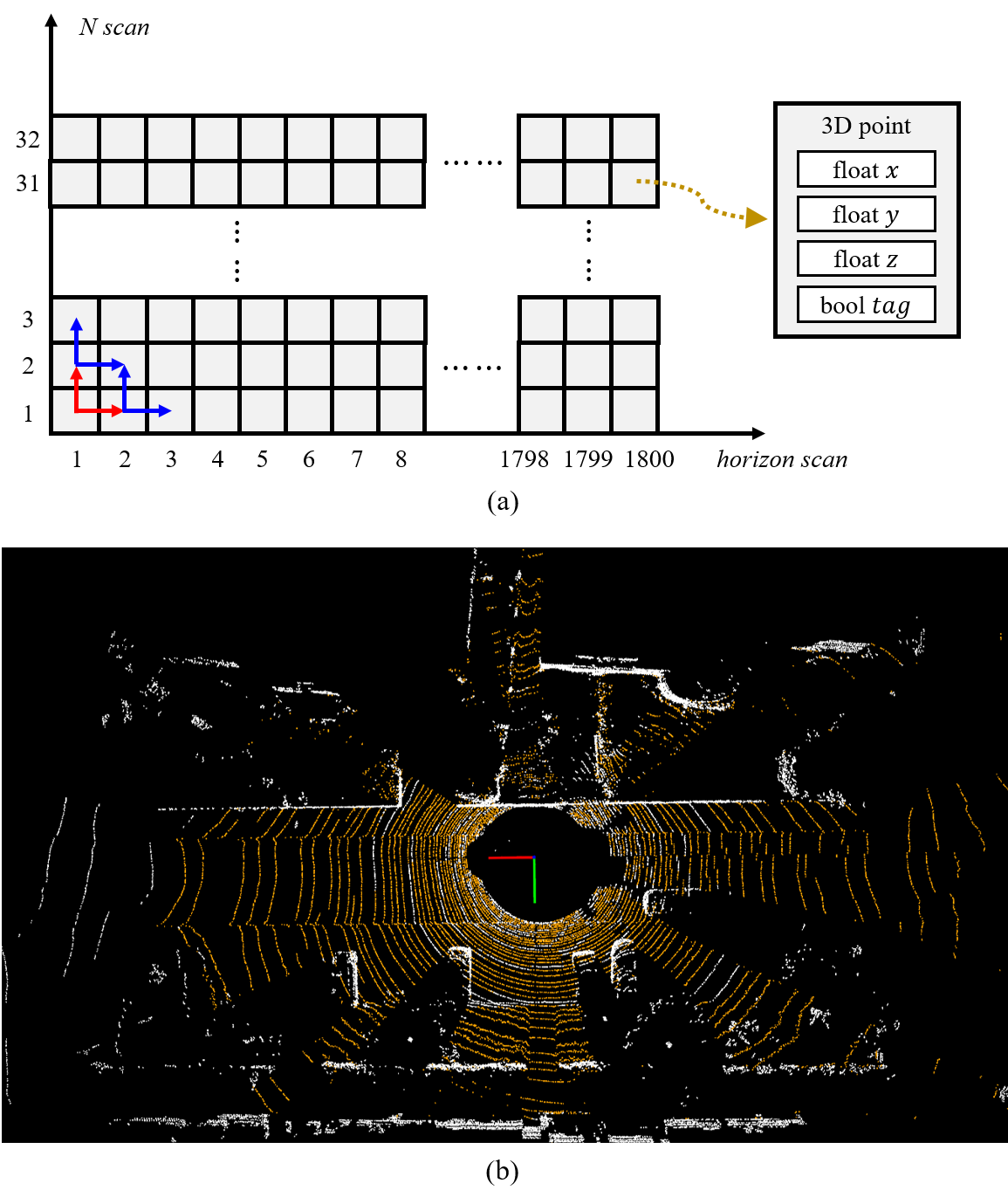}
		\caption{(a) Illustration of binarized label construction. The 3D points of current sweep are positioned in a range image, and a recursive 2D connected component method is utilized to identify ground points. (b) Visualization of segmented ground points from current input sweep. The orange points represent ground points and the white points represent non-ground points.}
		\label{fig3}
	\end{center}
\end{figure}

\subsection{Binarized Label Construction}
\label{Binarized Label Construction}

Before dynamic point identification, we first construct binarized descriptors, i.e., ground label and non-ground label, for each 3D point of current sweep. We utilize a fast 2D connected component method \cite{himmelsbach2010fast}, which is the same as LeGO-LOAM \cite{shan2018lego}, to separate ground points from current input sweep with very low computational cost. By default, the LiDAR is mounted horizontally on the vehicle, and the LiDAR's $z$-axis is perpendicular to the ground. If the LiDAR is equipped at an inclined angle, external parameters can be introduced to ensure the $z$-axis remains perpendicular. After changing the $z$-axis to perpendicular, we configure the range image $img$ \cite{nguyen2007comparison} with the number of LiDAR lines ($N$ $scan$ in Fig. \ref{fig3} (a)) as the vertical axis and the horizontal resolution ($horizon$ $scan$ in Fig. \ref{fig3} (a)) as the horizontal axis, and position the 3D points of current sweep in their corresponding locations within the range image according to their horizontal and vertical line indices. In the range image, each point is equipped with $x$, $y$, $z$ coordinates in $(\cdot)^l$, as well as a Boolean variable $tag$, which is used to denote whether the point is a ground point. Initially, all $tag$ are set to false. We set the $tag$ of point at $img(1,1)$ to $true$. Then following the red arrows indicated in Fig. \ref{fig3} (a), we calculate the pitch angles of two 3D points at the adjacent positions $img(1,2)$ and $img(2,1)$ to that at $img(1,1)$ respectively. Then the pitch angle of $img(2,1)$ to $img(1,1)$ can be calculated as:
\begin{equation}
	\begin{gathered}
		diff_x =img(2, 1).x-img(1, 1).x \\
		diff_y =img(2, 1).y-img(1, 1).y \\
		diff_z =img(2, 1).z-img(1, 1).z \\
		pitch = arctan \left(diff_z, sqrt\left(diff_x{ }^2+diff_y{ }^2\right)\right)
	\end{gathered}
\end{equation}
If the calculated pitch angle is less than a certain threshold (e.g., 5 degree in our system), $img(2,1).tag$ is set as $true$, which means the corresponding 3D point is determined as ground point. Similarily, we can calculate the pitch angle of $img(1,2)$ to $img(1,1)$, and set value for $img(1,2).tag$. If the 3D point located at $img(1,2)$ or $img(2,1)$ is determined as ground point, we follow the blue arrows indicated in Fig. \ref{fig3} (a) to indentify other points adjacent to them. The entire process is executed recursively, continuing until all pixels in the range image have been visited or the recursive exit condition is met. The visualization of segmented ground points is illustrated in Fig. \ref{fig3} (b), where the orange points are labeled as “ground points” and the white points are labeled as “non-ground points”.

\subsection{Background Separation}
\label{Background Separation}

In the process of executing label consistency detection, it is necessary to find the nearest neighbors for each point of current sweep. Points that are close to the vehicle platform can reliably find their nearest neighbors, whereas points that are farther may fail due to the incomplete reconstruction of their locations. In outdoor driving scenarios (excluding extreme occlusion), the map structures within a 30 meter radius around the vehicle platform are usually already reconstructed. Therefore, we set a empirical threshold of 30 meters, and define points within 30 meters of the vehicle platform as fore-points and those beyond 30 meters as back-points. For fore-points and back-points, we employ determinations that are specifically tailored to their characteristics.

\subsection{Voxel-Location-Based Nearest Neighbor Search}
\label{Voxel-Location-Based Nearest Neighbor Search}

\begin{figure}
	\begin{center}
		\includegraphics[scale=0.28]{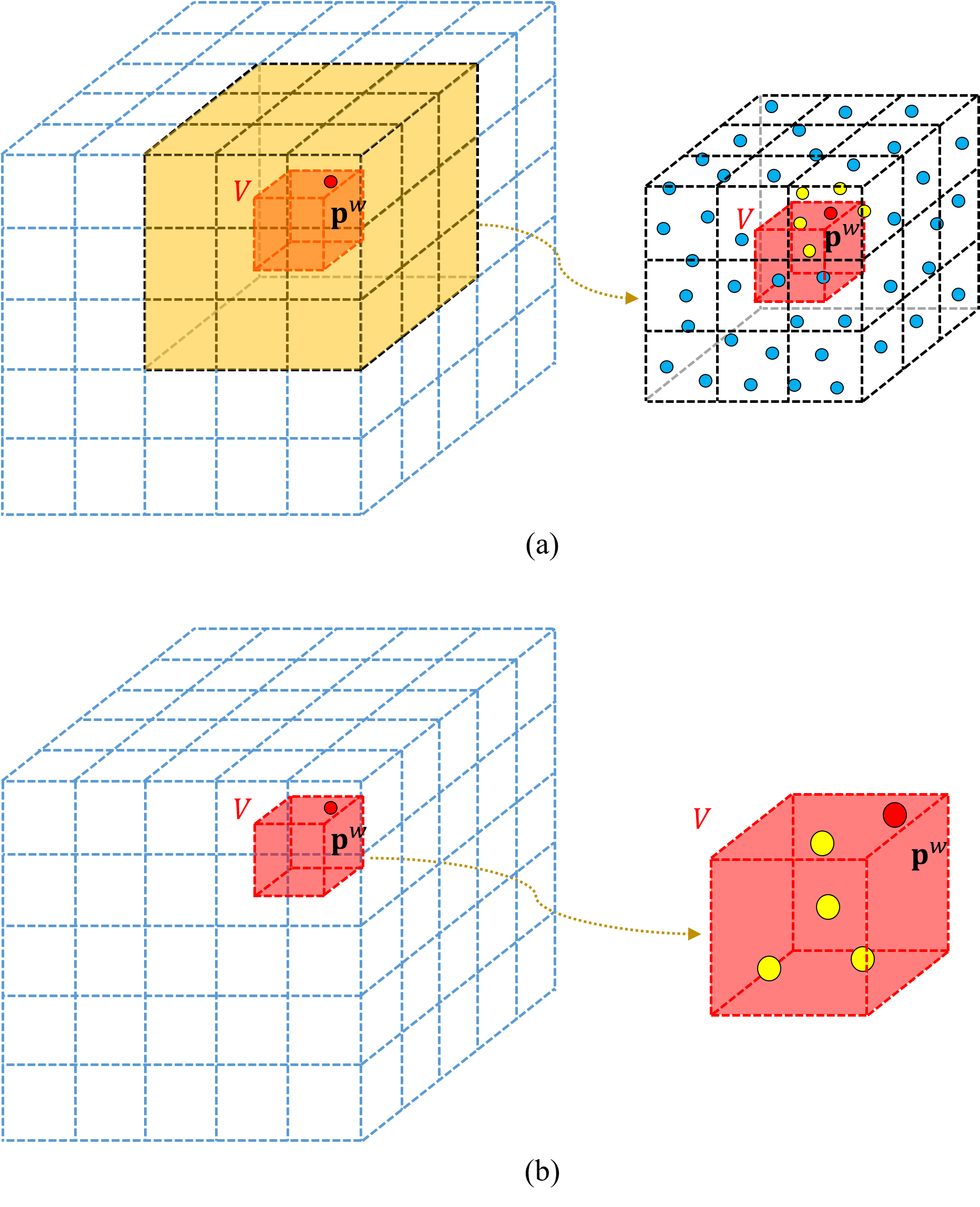}
		\caption{(a) Illustration of the conventional 8-nearest neighbor search. All points within the 8-nearest neighbor voxels are compared with the distance to $\mathbf{p}^w$, and the 20 closest points are selected as the nearest neighbors (simply represented by 5 yellow points). (b) Illustration of the voxel-location-based nearest neighbor search. The voxel $V$ to which $\mathbf{p}^w$ belongs is directly located, and all points in $V$ (no more than 20) are considered as nearest neighbors (simply represented by 5 yellow points).}
		\label{fig8}
	\end{center}
\end{figure}

For a specific point $\mathbf{p}^w$ with the “non-ground point” label, to determine whether its label is consistent with surrounding points in the global map, we first need to search for the nearest neighbors of $\mathbf{p}^w$. An intuitive alternative is to utilize the 8-nearest neighbor search which is the same as the nearest neighbor search method for point-to-plane distance computation (as illustrated in Fig. \ref{fig8} (a)) in LIO. Specially, we locate the voxel $V$ to which $\mathbf{p}^w$ belongs and the 8 voxels adjacent to $V$, and set all points in these voxels as candidate points. Subsequently, the 20 nearest points of $\mathbf{p}^w$ are identified from 9 candidate voxels by comparing the magnitudes of Euclidean distances to $\mathbf{p}^w$.

However, calculating the Euclidean distance of each candidate point to $\mathbf{p}^w$ is an extremely time-consuming process. When LIO builds the point-to-plane distance residuals, in order to ensure that the fitted plane can reflect as much of the geometric information around $\mathbf{p}^w$ as possible, we have to utilize the conventional 8-nearest neighbor search. Fortunately, only 600 point-to-plane distance residuals are required for estimating the pose of each sweep in our system, so the total computational overhead is acceptable. However, in order to ensure that the final output map does not contain dynamic points, it is necessary to determine each point of the current sweep, which requires searching the nearest neighbors for each point from the global map. A single sweep of one 32-line LiDAR can yield more than 50,000 points, making the conventional 8-nearest neighbor search inapplicable here.

In the proposed label consistency detection method, we qualitatively assess whether the non-ground point $\mathbf{p}^w$ is a dynamic point by comparing its label with those of its surround points. Since the surround points are not involved in quantitative calculations, it is not necessary to strictly satisfy the concept of nearest neighbors. Instead, an approximate approach, i.e., voxel-location-based nearest neighbor search, can be adopted to significantly reduce the computational cost. As illustrated in Fig. \ref{fig8} (b), we locate the voxel $V$ to which $\mathbf{p}^w$ belongs and consider the other points within $V$ as the approximate nearest neighbors. Since the voxel map has a query operation with a computational complexity of $O(1)$, the entire nearest neighbor search process is extremely fast. In addition, the computational overhead associated with Euclidean distance is also saved. The voxel-location-based nearest neighbor search plays a crucial role in ensuring the low computational cost of label consistency detection, as evidenced by the results of the ablation study, which are documented in Sec. \ref{Ablation Study of Nearest Neighbor Search}.

\begin{figure}
	\begin{center}
		\includegraphics[scale=0.45]{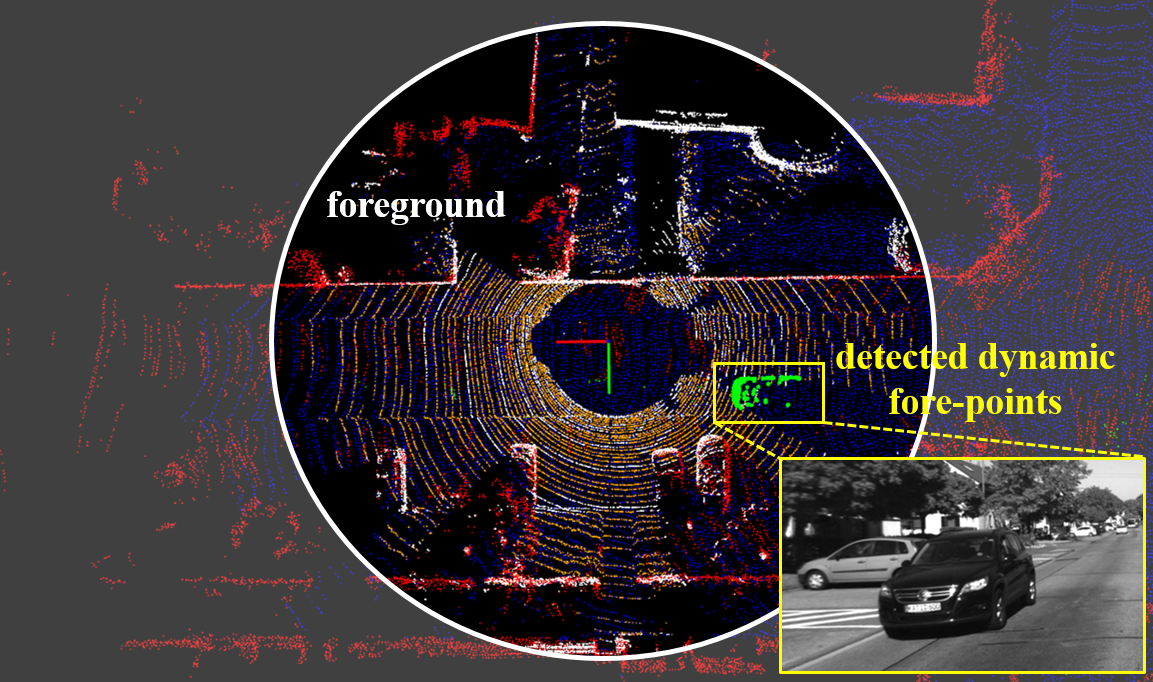}
		\caption{Visualization of dynamic point determination results for fore-points. The areas obscured in white are the background regions, while those that remain visible are the foreground regions. The orange and white points represent the ground and non-ground points of current sweep, and the blue and red points correspond to the ground and non-ground points of global map. The green points are the identified dynamic points.}
		\label{fig5}
	\end{center}
\end{figure}

\subsection{Dynamic Point Determination}
\label{Dynamic Point Determination}

In Sec. \ref{Background Separation}, we categorize the points of current sweep into fore-points and back-points based on their distance from the vehicle platform. For dynamic point determination of fore-points and back-points, we employ the following two distinct modes.

\textbf{Mode for fore-points.} If the number of nearest neighbors is below a certain threshold (5 in our system), it indicates that the location of $\mathbf{p}^w$ was originally unoccupied, and thus $\mathbf{p}^w$ is classified as a dynamic point. If the number of nearest neighbors is sufficiently large (greater than 5), we calculate the proportion of non-ground points among all nearest neighbors. If this proportion is sufficiently low (less than 30$\%$), $\mathbf{p}^w$ is classified as a static point and added to both the tracking-map and the output map. Conversely, if the proportion is larger than 30$\%$, $\mathbf{p}^w$ is classified as a dynamic point and excluded in the map. Inevitably, this determination criteria potentially results in the erroneous removal of some static points in close proximity to the ground. Nonetheless, the points most commonly affected by this misfiltration are situated at the transition between walls and the ground. Despite the possibility of such misfiltration, the overall geometric integrity of the scene remains intact, and it does not affect the performance of the LIO system. The visualization of dynamic point determination results for fore-points is shown in Fig. \ref{fig5}.

\begin{figure}
	\begin{center}
		\includegraphics[scale=0.45]{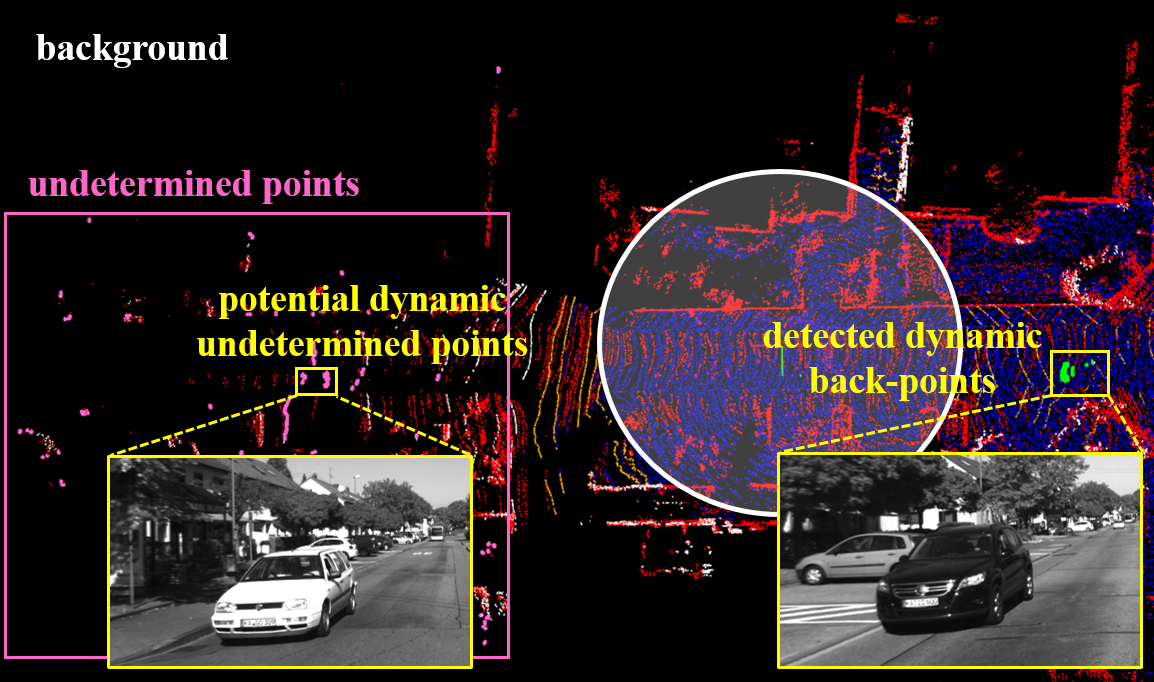}
		\caption{Visualization of dynamic point determination results for back-points. The areas obscured in white are the foreground regions, while those that remain visible are the background regions. The orange and white points represent the ground and non-ground points of current sweep, and the blue and red points correspond to the ground and non-ground points of global map. The green points are the identified dynamic points, and the pink points are the undetermined points, which need to be re-determined after the map arround them has been reconstructed.}
		\label{fig6}
	\end{center}
\end{figure}

\textbf{Mode for back-points.} If the number of nearest neighbors is below a certain threshold (5 in our system), we cannot identify the back-point as a dynamic point, because it is possible that the location has not yet been reconstructed, preventing the obtainment of the nearest neighbors. Such points are labeled as undetermined-points, and a determination will be made once the vehicle platform continues to move and the geometric structures of the locations of these points are recovered. To ensure that newly acquired point clouds can be properly registered during state estimation, it is necessary to incorporate the undetermined-points into the tracking-map. This will not significantly affect the accuracy of state estimation, as even if there are dynamic objects among the back-points, the number of LiDAR points scanned onto them is very sparse. As for the final output map, it is imperative to ensure that it contains as few dynamic points as possible, hence the determination for undetermined-points will be conducted subsequently. When the number of nearest neighbors is sufficiently large (greater than 5), the processing approach is the same as that for fore-points, and the static points are added to both the tracking-map and the output map. The visualization of dynamic point determination results for back-points is shown in Fig. \ref{fig6}.

\subsection{Undetermined-Point Re-Determination}
\label{Undetermined-Point Re-Determination}

\begin{figure}
	\begin{center}
		\includegraphics[scale=0.45]{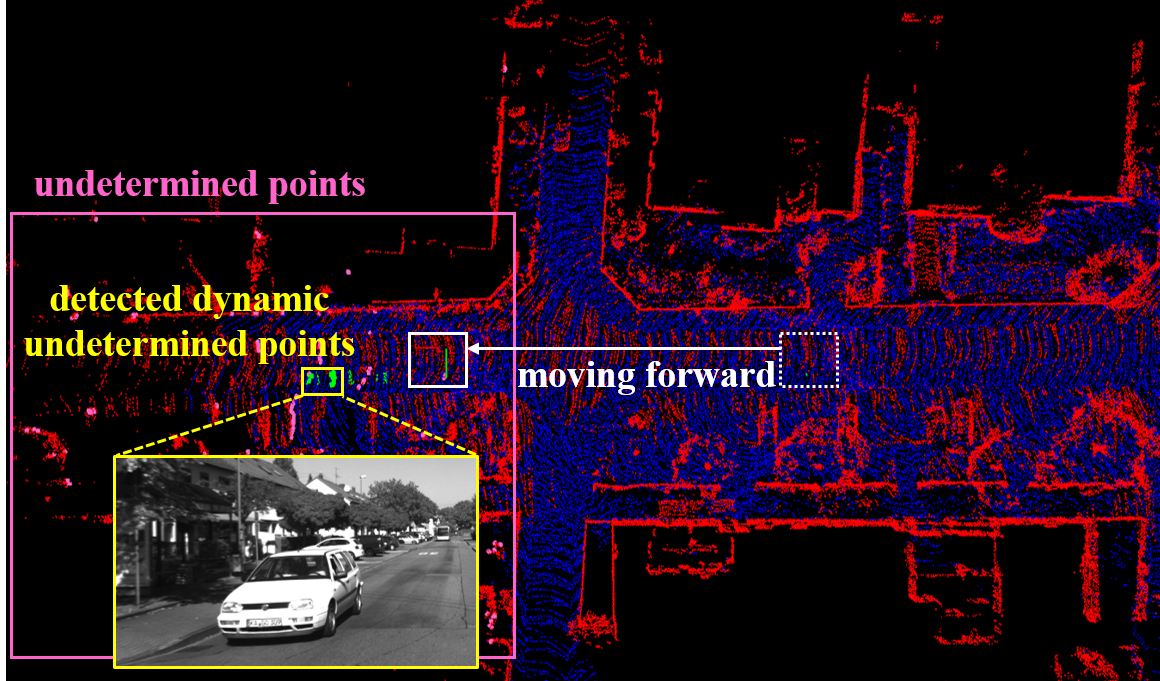}
		\caption{Visualization of dynamic point determination results for undetermined-points. The blue and red points correspond to the ground and non-ground points of global map. As the vehicle platform continues to move forward, the geometric structure information of previously unreconstructed positions of the global map is recovered. Then the undetermined-points can be re-determined by label consistency detection.}
		\label{fig7}
	\end{center}
\end{figure}

As mentioned in Sec. \ref{Dynamic Point Determination}, some back-points may fail to find nearest neighbors due to the incomplete reconstruction of their locations. For such points, we label them as undetermined points and place them in a container. As the vehicle platform continues to move forward, the geometric structure informations of previously unreconstructed positions are recovered (as shown in Fig. \ref{fig7}). Then we can make re-determination for those undetermined-points. When a point $\mathbf{p}_u^w$ in the undetermined point container is close to the current position of the vehicle platform (less than 30 meter), it is highly likely that the geometric structure information around $\mathbf{p}_u^w$ has been reconstructed. We can then determine whether $\mathbf{p}_u^w$ is a dynamic point. If the number of nearest neighbors is below a certain threshold (5 in our system), it suggests that the location of $\mathbf{p}_u^w$ was originally unoccupied, leading to the classification of $\mathbf{p}_u^w$ as a dynamic point. If the number of nearest neighbors is larger than the threshold of 5, we calculate the proportion of non-ground points among all nearest neighbors. If this proportion is sufficiently low (less than 30$\%$), it is classified as a static point and added to the output map. On the contrary, if the proportion is not smaller than the threshold of 30$\%$, it is classified as a dynamic point and would not be included in output map. If an undetermined-point is more than 30 meters away from the vehicle platform's position for 10 consecutive sweeps, it is likely to be a sparse background point at far distance. Thus, we directly classify it as a static point and add it to the output map.

\section{Our System Dynamic-LIO}
\label{Our System Dynamic-LIO}

\subsection{System Overview}
\label{System Overview}

\begin{figure}
	\begin{center}
		\includegraphics[scale=0.41]{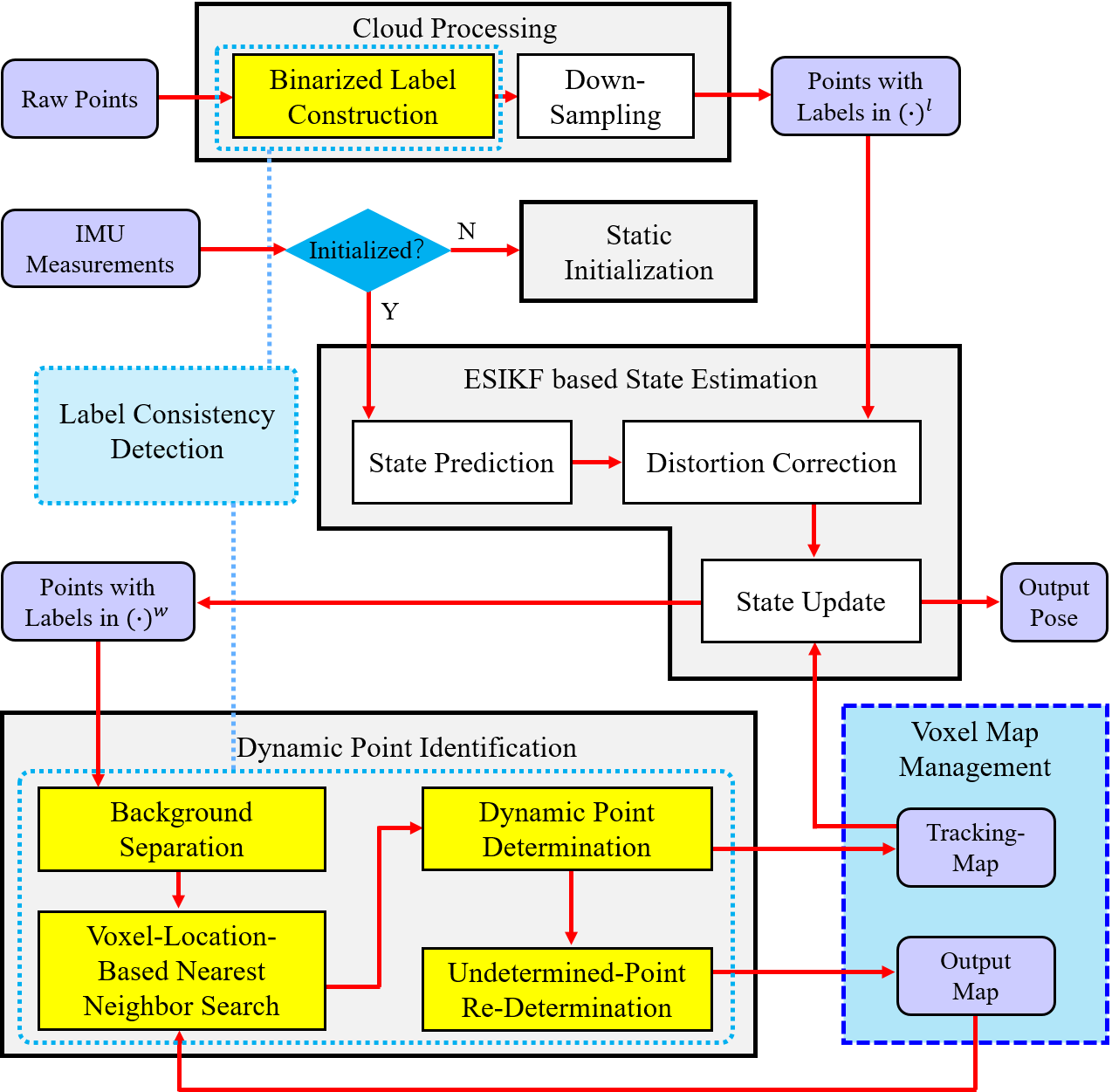}
		\caption{Overview of our Dynamic-LIO which consists of four main modules: cloud processing, static initialization, ESIKF based state estimation and dynamic point identification. The yellow rectangles indicate the operations of our system that are related to label consistency detection.}
		\label{fig2}
	\end{center}
\end{figure}

Fig. \ref{fig2} illustrates the framework of our self-developed system Dynamic-LIO which consists of four main modules: cloud processing, static initialization, ESIKF based state estimation and dynamic point identification. The cloud processing module constructs binarized label (i.e., ground label or non-ground label) for each 3D point of current sweep. Subsequently, it performs spatial down-sampling to ensure uniform density of current point cloud. The static initialization module \cite{geneva2020openvins} utilizes the IMU measurements to estimate some state parameters such as gravitational acceleration, accelerometer bias, gyroscope bias and initial velocity. The ESIKF based state estimation module estimates the state of current sweep, and utilize the estimated pose to transform all points of current sweep from $(\cdot)^l$ to $(\cdot)^w$. The dynamic point identification module identifies dynamic points from current input point cloud data, to ensure that the global map contains only static points. The yellow rectangles indicate the specific locations of five steps of label consistency detection within the overall system framework.

\subsection{Down-Sampling}
\label{Down-Sampling}

To alleviate the substantial computational load caused by the overwhelming number of 3D points collected by a LiDAR in a single sweep, we implement a decimation strategy on point cloud. Initially, a uniform subsampling approach is applied to retain a single point for every set of four. Subsequently, we integrate the uniformly subsampled points into a voxel grid defined by a $0.5\times0.5\times0.5$ (unit: m) resolution, ensuring that each voxel is occupied by a solitary point.

It is worth noting that the down-sampling of current input sweep must be performed after binarized label construction. The reason is that down-sampling can disrupt the adjacency relationships of 3D points in range images, which can affect the execution of binarized label construction.

\subsection{Static Initialization}
\label{Static Initialization}

In our Dynamic-LIO, a static initialization procedure is employed to estimate essential parameters, including initial poses, initial velocities, gravitational forces, as well as biases in both accelerometer and gyroscope measurements. For a detailed elucidation of the methodology, please refer to reference \cite{geneva2020openvins}.

\subsection{ESIKF based State Estimation}
\label{ESIKF based State Estimation}

We adapt the error state iterated Kalman filter (ESIKF) to perform state estimation, which is the same as Fast-LIO2 \cite{xu2022fast}. Fast-LIO2 utilized their own self-developed toolbox IKFoM \cite{he2021kalman} for the implementation of on-manifold Kalman filter, and we utilized the more widely recognized Eigen3 library \cite{eugene2011eigen3} to implement this function. We have documented the entire execution process of our ESIKF in Appendices for readers to refer the details of implementation.

It is worth mentioning that during state estimation, we need to find the nearest neighbors for randomly selected 600 points of current sweep from the global map, and fit a plane with these nearest neighbors to construct point-to-plane distance constraints. To ensure that the final fitted plane as accurately as possible reflect the surrounding geometric information, we still use the conventional 8-nearest neighbor search method here. Additionally, we search for nearest neighbors based on the tracking-map. Compared to the output map, the tracking-map is more conducive to the accurate and robust running of LIO. Because the tracking-map retains enough geometric details while removing out the vast majority of dynamic points.

\section{Experiments}
\label{Experiments}

We evaluate the overall performance of our method on six autonomous driving scenario datasets: $semantic$-$kitti$ \cite{behley2019semantickitti}, $ulhk$-$CA$ \cite{wen2020urbanloco}, $urban$-$Nav$ \cite{hsu2021urbannav}, $nclt$ \cite{carlevaris2016university}, $utbm$ \cite{yan2020eu} and $ulhk$-$HK$ \cite{wen2020urbanloco}. Among them, $semantic$-$kitti$, $ulhk$-$CA$ and $urban$-$Nav$ are three public datasets collected at dynamic scenes, $nclt$, $utbm$ and $ulhk$-$HK$ are three public datasets collected at static scenes. $semantic$-$kitti$ is collected by a 64-line Velodyne LiDAR and each LiDAR point has its unique semantic label. Thus, $semantic$-$kitti$ is used to evaluate the preservation rate (PR) and rejection rate (RR) of proposed label consistency based dynamic point detection and removal method. $ulhk$-$CA$ is collected by a 32-line Robosense LiDAR and IMU, and $urban$-$Nav$ is collected by a 32-line Velodyne LiDAR and IMU. These two datasets are used to evaluate the improvement of dynamic point detection and removal method on pose estimation in terms of absolute trajectory error (ATE). Both $nclt$, $utbm$ and $ulhk$-$HK$ are collected by a 32-line Velodyne LiDAR and IMU. These three datasets are used to show the outstanding performance of our self-developed LIO system, and demonstrate that the proposed dynamic point detection and removal method has no negative effect on the accuracy of LIO in static scenes. Details of all the 24 sequences used in this section, including name, duration, and whether they contain dynamic objects, are listed in Table \ref{table31}. A consumer-level computer equipped with an Intel Core i7-11700 and 32 GB RAM is used for all experiments.

\begin{table}[]
	\begin{center}
	\caption{Datasets of All Sequences for Evaluation}
	\label{table31}
	\begin{tabular}{cccc}
		\hline
		& Name              & \begin{tabular}[c]{@{}c@{}}Duration\\ (min:sec)\end{tabular} & \begin{tabular}[c]{@{}c@{}}Whether Dynamic \\ Objects are Included\end{tabular} \\ \hline
		$kitti\_1$ & semantic-kitti-00 & 00:14                                                        & Yes                                                                            \\
		$kitti\_2$ & semantic-kitti-01 & 00:10                                                        & Yes                                                                            \\
		$kitti\_3$ & semantic-kitti-02 & 00:09                                                        & Yes                                                                            \\
		$kitti\_4$ & semantic-kitti-05 & 00:33                                                        & Yes                                                                            \\
		$kitti\_5$ & semantic-kitti-07 & 00:20                                                        & Yes                                                                            \\
		$ulhk\_1$  & CA-MarktStreet    & 24:35                                                        & Yes                                                                            \\
		$ulhk\_2$  & CA-RussianHill    & 26:57                                                        & Yes                                                                            \\
		$urban\_1$ & TST               & 13:05                                                        & Yes                                                                            \\
		$urban\_2$ & Whampoa           & 25:37                                                        & Yes                                                                            \\
		$nclt\_1$  & 2012-01-08        & 92:16                                                        & No                                                                             \\
		$nclt\_2$  & 2012-02-02        & 98:37                                                        & No                                                                             \\
		$nclt\_3$  & 2012-02-04        & 77:39                                                        & No                                                                             \\
		$nclt\_4$  & 2012-05-11        & 83:36                                                        & No                                                                             \\
		$nclt\_5$  & 2012-05-26        & 97:23                                                        & No                                                                             \\
		$nclt\_6$  & 2012-06-15        & 55:10                                                        & No                                                                             \\
		$nclt\_7$  & 2012-08-04        & 79:27                                                        & No                                                                             \\
		$nclt\_8$  & 2012-09-28        & 76:40                                                        & No                                                                             \\
		$utbm\_1$  & 2018-07-19        & 15:26                                                        & No                                                                             \\
		$utbm\_2$  & 2019-01-31        & 16:00                                                        & No                                                                             \\
		$utbm\_3$  & 2019-04-18        & 11:59                                                        & No                                                                             \\
		$utbm\_4$  & 2018-07-20        & 16:45                                                        & No                                                                             \\
		$utbm\_5$  & 2018-07-13        & 16:59                                                        & No                                                                             \\
		$ulhk\_3$  & HK-2019-01-17     & 05:18                                                         & No                                                                             \\
		$ulhk\_4$  & HK-2019-04-26-1   & 02:30                                                         & No                                                                             \\ \hline
	\end{tabular}
	\end{center}
\end{table}

\subsection{PR and RR Comparison with the State-of-the-Arts}
\label{PR and RR Comparison with the State-of-the-Arts}

We compare our label consistency based dynamic point detection method with three state-of-the-art 3D point-based dynamic point detection methods, i.e., Removert \cite{kim2020remove}, Erasor \cite{lim2021erasor} and Dynamic Filter \cite{fan2022dynamicfilter}, on $semantic$-$kitti$ dataset \cite{behley2019semantickitti}. Among them, Removert and Erasor are offline methods which need the pre-built map as input, Dynamic Filter and our method are online methods which do not rely on any prior information.

\begin{table}[]
	\begin{center}
		\caption{PR Comparison with State-of-the-Art Methods (unit: $\%$)}
		\label{table1}
		\begin{tabular}{c|cc|cc}
			\hline
			\multirow{2}{*}{} & \multicolumn{2}{c|}{offline} & \multicolumn{2}{c}{online} \\ \cline{2-5} 
			& Removert       & Erasor      & Dynamic Filter   & Ours    \\ \hline
			$kitti\_1$ & 86.83          & \textbf{93.98}       & 90.07            & \textbf{90.36}   \\
			$kitti\_2$ & \textbf{95.82}          & 91.49       & 87.95            & \textbf{88.43}   \\
			$kitti\_3$ & 83.29          & \textbf{87.73}       & 88.02            & \textbf{88.25}   \\
			$kitti\_4$ & 88.17          & \textbf{88.73}       & 90.17            & \textbf{90.31}   \\
			$kitti\_5$ & 82.04          & \textbf{90.62}       & 87.94            & \textbf{89.28}   \\ \hline
		\end{tabular}
	\end{center}
\end{table}

\begin{table}[]
	\begin{center}
		\caption{RR Comparison with State-of-the-Art Methods (unit: $\%$)}
		\label{table2}
		\begin{tabular}{c|cc|cc}
			\hline
			\multirow{2}{*}{} & \multicolumn{2}{c|}{offline} & \multicolumn{2}{c}{online} \\ \cline{2-5} 
			& Removert       & Erasor      & Dynamic Filter   & Ours    \\ \hline
			$kitti\_1$ & 90.62          & \textbf{97.08}       & \textbf{91.09}            & 90.73   \\
			$kitti\_2$ & 57.08          & \textbf{95.38}       & 87.69            & \textbf{88.41}   \\
			$kitti\_3$ & 88.37          & \textbf{97.01}       & 86.10            & \textbf{86.22}   \\
			$kitti\_4$ & 79.98          & \textbf{98.26}       & 84.65            & \textbf{85.84}   \\
			$kitti\_5$ & 95.50          & \textbf{99.27}       & 86.80            & \textbf{87.34}   \\ \hline
		\end{tabular}
	\end{center}
\end{table}

Results in Table \ref{table1} and Table \ref{table2} demonstrate that our method outperforms Dynamic Filter for almost all sequences in terms of higher PR and RR. Although Dynamic Filter achieves higher RR than our method on sequence $kitti\_1$, our result is very close to theirs, with only a 0.36$\%$ difference.

\subsection{ATE Comparison with the State-of-the-Arts}
\label{ATE Comparison with the State-of-the-Arts}

We compare our Dynamic-LIO with two state-of-the-art LIO systems for dynamic scenes, i.e., RF-LIO \cite{qian2022rf} and ID-LIO \cite{wu2023lidar}, on $ulhk$-$CA$ \cite{wen2020urbanloco} and $urban$-$Nav$ \cite{hsu2021urbannav} datasets. Both RF-LIO, ID-LIO have loop detection module, and use GTSAM \cite{kaess2012isam2} to optimize the factor graph. Thus, we also add the same loop detection module and global optimization module to Dynamic-LIO when comparing with them. Both results of RF-LIO and ID-LIO are recorded from their literatures because they have not released the code. The selected four sequences both encompass highly dynamic scenarios and can effectively evaluate the performance of LIO systems in dynamic scenes.

\begin{table}[]
	\begin{center}
		\caption{RMSE of ATE Comparison with State-of-the-Art Methods on Datasets of Dynamic Scenes (unit: m)}
		\label{table3}
		\begin{threeparttable}
			\begin{tabular}{p{1.5cm}<{\centering}|p{1.5cm}<{\centering}p{1.5cm}<{\centering}|p{1.5cm}<{\centering}}
				\hline
				& RF-LIO & ID-LIO & Ours           \\ \hline
				$ulhk\_1$   & 15.89                                             & 28.02                                             & \textbf{12.96} \\
				$ulhk\_2$   & 12.17                                             & 15.34                                             & \textbf{4.84}  \\
				$urban\_1$    & -                                                 & \textbf{1.06}                                     & 3.90           \\
				$urban\_2$ & -                                                 & \textbf{3.45}                                     & 6.66           \\ \hline
			\end{tabular}
		\end{threeparttable}
		\begin{tablenotes}
			\footnotesize
			\item[] \textbf{Denotations}: "-" means the corresponding value is not available.
		\end{tablenotes}
	\end{center}
\end{table}

Results in Table \ref{table3} demonstrate that the accuracy of our Dynamic-LIO is superior to that of RF-LIO and ID-LIO on $ulhk\_1$ and $ulhk\_2$. Since RF-LIO is neither open-sourced nor tested on the $urban$-$Nav$ dataset, we are unable to obtain its results on sequence $urban\_1$ and $urban\_2$. Although ID-LIO achieves smaller ATE than our system on $urban$-$Nav$ dataset, our act of open-sourcing the code better substantiates the reproducibility of our results.

We also compare our Dynamic-LIO with six state-of-the-art LIO systems for static scenes, i.e., LiLi-OM \cite{li2021towards}, LIO-SAM \cite{shan2020lio}, Fast-LIO2 \cite{xu2022fast}, DLIO \cite{chen2023direct}, IG-LIO \cite{chen2024ig} and Point-LIO \cite{he2023point}, on $nclt$ \cite{carlevaris2016university}, $utbm$ \cite{yan2020eu} and $ulhk$-$HK$ \cite{wen2020urbanloco} datasets, to demonstrate that our Dynamic-LIO still has excellent pose estimation performance in static scenes. The six methods we compared have released corresponding code, thus we obtain the results of them based on the source code provided by the authors.

\begin{table}[]
	\begin{center}
		\caption{RMSE of ATE Comparison of State-of-the-art Methods on Datasets of Static Scenes (Unit: m)}
		\label{table32}
		\begin{threeparttable}
			\begin{tabular}{c|p{0.6cm}<{\centering}p{0.6cm}<{\centering}p{0.55cm}<{\centering}p{0.55cm}<{\centering}p{0.55cm}<{\centering}p{0.55cm}<{\centering}|p{0.5cm}<{\centering}}
				\hline
				& \begin{tabular}[c]{@{}c@{}}LiLi-\\ OM\end{tabular} & \begin{tabular}[c]{@{}c@{}}LIO-\\ SAM\end{tabular} & \begin{tabular}[c]{@{}c@{}}Fast-\\ LIO2\end{tabular} & DLIO           & \begin{tabular}[c]{@{}c@{}}IG-\\ LIO\end{tabular} & \begin{tabular}[c]{@{}c@{}}Point-\\ LIO\end{tabular} & Ours           \\ \hline
				$nclt\_1$  & 50.71                                              & 1.85                                               & 3.57                                                 & 3.27           & 1.85                                              & 2.55 & \textbf{1.54}  \\
				$nclt\_2$  & 91.86                                              & 7.18                                               & 2.00                                                 & 1.80           & \textbf{1.72}                                     & 2.45 & 1.74           \\
				$nclt\_3$  & 92.93                                              & \textbf{2.16}                                      & 2.77                                                 & 5.35           & 2.92                                              & 5.31 & 2.23           \\
				$nclt\_4$  & 185.24                                             & $\times$                                                  & 2.46                                                 & 3.14           & 1.84                                              & 11.24 & \textbf{1.67}  \\
				$nclt\_5$  & 141.83                                             & $\times$                                                  & 2.60                                                 & 12.44          & \textbf{2.12}                                              & 14.89 & 2.24  \\
				$nclt\_6$  & 50.42                                              & 2.97                                               & 2.37                                                 & 2.98           & \textbf{1.82}                                     & 4.39 & 2.05           \\
				$nclt\_7$  & 137.05                                             & 2.26                                               & 2.59                                                 & 7.84           & 2.40                                     & 16.28 & \textbf{2.13}           \\
				$nclt\_8$ & $\times$                                                  & $\times$                                                  & 2.65                                                 & 7.72           & 1.72                                              & 16.22 & \textbf{1.66}  \\
				$utbm\_1$  & 67.16                                              & -                                                  & 15.13                                                & 14.25          & 17.37                                             & 22.71 & \textbf{13.92}  \\
				$utbm\_2$  & 38.17                                              & -                                                  & 21.21                                                & \textbf{13.85} & 21.27                                             & 23.02 & 16.09          \\
				$utbm\_3$  & 10.70                                              & -                                                  & 10.81                                                & 55.28          & 13.75                                             & 13.81 & \textbf{9.10}  \\
				$utbm\_4$  & 70.98                                              & -                                                  & 15.20                                                & 18.05          & 16.44                                             & 21.76 & \textbf{9.63} \\
				$utbm\_5$  & 62.57                                              & -                                                  & 13.24                                                & 14.95          & $\times$                                          & 19.88  & \textbf{9.63}  \\
				$ulhk\_3$  & $\times$                                                  & 1.68                                               & 1.20                                                 & 2.44           & 1.15                                     & 1.07 & \textbf{1.03}           \\
				$ulhk\_4$  & 3.11                                      & 3.13                                               & 3.24                                                 & $\times$              & 3.31                                    & \textbf{2.82}      & 3.08           \\ \hline
			\end{tabular}
		\end{threeparttable}
		\begin{tablenotes}
			\footnotesize
			\item[] \textbf{Denotations}: “$\times$” means the system fails to run entirely on the corresponding sequence, and “-” means the corresponding value is not available.
		\end{tablenotes}
	\end{center}
\end{table}

\begin{table}[]
	\begin{center}
		\caption{Impact of Undetermined-Points on PR (unit: $\%$)}
		\label{table4}
		\begin{tabular}{p{1.5cm}<{\centering}|p{3.0cm}<{\centering}p{1.5cm}<{\centering}}
			\hline
			& \begin{tabular}[c]{@{}c@{}}Ours w/o Considering\\ Undetermined-Points\end{tabular} & Ours           \\ \hline
			$kitti\_1$ & 90.09                                                                              & \textbf{90.36} \\
			$kitti\_2$ & 88.17                                                                              & \textbf{88.43} \\
			$kitti\_3$ & 87.59                                                                              & \textbf{88.25} \\
			$kitti\_4$ & 89.31                                                                              & \textbf{90.31} \\
			$kitti\_5$ & 88.43                                                                              & \textbf{89.28} \\ \hline
		\end{tabular}
	\end{center}
\end{table}

\begin{table}[]
	\begin{center}
		\caption{Impact of Undetermined-Points on RR (unit: $\%$)}
		\label{table5}
		\begin{tabular}{p{1.5cm}<{\centering}|p{3.0cm}<{\centering}p{1.5cm}<{\centering}}
			\hline
			& \begin{tabular}[c]{@{}c@{}}Ours w/o Considering\\ Undetermined-Points\end{tabular} & Ours           \\ \hline
			$kitti\_1$ & 90.05                                                                              & \textbf{90.73} \\
			$kitti\_2$ & 87.86                                                                              & \textbf{88.41} \\
			$kitti\_3$ & \textbf{86.22}                                                                              & \textbf{86.22} \\
			$kitti\_4$ & 84.08                                                                              & \textbf{85.84} \\
			$kitti\_5$ & 87.30                                                                              & \textbf{87.34} \\ \hline
		\end{tabular}
	\end{center}
\end{table}

\begin{table}[]
	\begin{center}
		\caption{Impact of Removing Dynamic Points on ATE (unit: m)}
		\label{table6}
		\begin{tabular}{c|c|cc}
			\hline
			&          & \begin{tabular}[c]{@{}c@{}}Ours w/o Removing\\ Dynamic Points\end{tabular} & Ours           \\ \hline
			\multirow{4}{*}{Dynamic Scenes} & $ulhk\_1$  & 13.98                                                                      & \textbf{12.96} \\
			& $ulhk\_2$  & 4.88                                                                       & \textbf{4.84}  \\
			& $urban\_1$ & 5.21                                                                       & \textbf{3.90}  \\
			& $urban\_2$ & 9.19                                                                       & \textbf{6.66}  \\ \hline
			\multirow{15}{*}{Static Scenes} & $nclt\_1$  & 1.60                                                                       & \textbf{1.54}  \\
			& $nclt\_2$  & 1.77                                                                       & \textbf{1.74}  \\
			& $nclt\_3$  & \textbf{2.23}                                                              & \textbf{2.23}  \\
			& $nclt\_4$  & 2.55                                                                       & \textbf{1.67}  \\
			& $nclt\_5$  & 2.44                                                                       & \textbf{2.24}  \\
			& $nclt\_6$  & \textbf{1.93}                                                              & 2.05           \\
			& $nclt\_7$  & \textbf{2.06}                                                              & 2.13           \\
			& $nclt\_8$  & 1.92                                                                       & \textbf{1.66}  \\
			& $utbm\_1$  & \textbf{13.76}                                                             & 13.92          \\
			& $utbm\_2$  & 17.05                                                                      & \textbf{16.09} \\
			& $utbm\_3$  & 9.66                                                                       & \textbf{9.10}  \\
			& $utbm\_4$  & 12.97                                                                      & \textbf{9.63}  \\
			& $utbm\_5$  & 9.74                                                                       & \textbf{9.63}  \\
			& $ulhk\_3$  & 1.04                                                                       & \textbf{1.03}  \\
			& $ulhk\_4$  & 3.31                                                                       & \textbf{3.08}  \\ \hline 
		\end{tabular}
	\end{center}
\end{table}

\begin{table}[]
	\begin{center}
		\caption{Time Consumption Comparison with State-of-the-Art Methods on Semantic-Kitti Dataset (unit: ms)}
		\label{table7}
		\begin{tabular}{p{1.5cm}<{\centering}|p{1.5cm}<{\centering}p{1.5cm}<{\centering}|p{1.5cm}<{\centering}}
			\hline
			& \begin{tabular}[c]{@{}c@{}}Dynamic Filter\\ (Front-End)\end{tabular} & RH-Map    & Ours            \\ \hline
			$kitti\_1$    & \multirow{5}{*}{55.71}                                               & 63.96     & \textbf{41.60}  \\
			$kitti\_2$    &                                                                      & 93.23     & \textbf{34.93}  \\
			$kitti\_3$    &                                                                      & 66.98     & \textbf{46.00}  \\
			$kitti\_4$    &                                                                      & 64.61     & \textbf{34.87}  \\
			$kitti\_5$    &                                                                      & 51.02     & \textbf{27.61}  \\ \hline
			CPU model   & i7-8559U                                                             & i7-12700H & i7-11700        \\
			clock speed & \textbf{2.7GHz}                                                               & \textbf{2.7Ghz}    & 2.5GHz \\ \hline
		\end{tabular}
	\end{center}
\end{table}

\begin{table}[]
	\begin{center}
		\caption{Time Consumption Comparison with State-of-the-Art LIO Systems on ULHK-CA and Urban-Nav Datasets (unit: ms)}
		\label{table8}
		\begin{threeparttable}
			\begin{tabular}{c|p{1.5cm}<{\centering}p{1.5cm}<{\centering}|p{1.5cm}<{\centering}}
				\hline
				& RF-LIO          & ID-LIO          & Ours     \\ \hline
				$ulhk\_1$     & 96              & 121             & \textbf{23.33}    \\
				$ulhk\_2$     & 121             & 100             & \textbf{21.50}    \\
				$urban\_1$    & -               & 96              & \textbf{16.46}    \\
				$urban\_2$    & -               & 99              & \textbf{16.41}    \\ \hline
				CPU model   & i5              & i7-10700K       & i7-11700 \\
				clock speed & 1.3$\sim$3.7GHz & \textbf{3.8GHz} & 2.5GHz   \\ \hline
			\end{tabular}
		\end{threeparttable}
		\begin{tablenotes}
			\footnotesize
			\item[] \textbf{Denotations}: “-” means the corresponding value is not available.
		\end{tablenotes}
	\end{center}
\end{table}

\begin{table*}[]
	\begin{center}
		\caption{Time Consumption of Each Module (unit: ms)}
		\label{table9}
		\begin{tabular}{c|cc|ccc|c}
			\hline
			\multirow{2}{*}{} & \multirow{2}{*}{Cloud Processing} & \multirow{2}{*}{State Estimation} & \multicolumn{3}{c|}{Label Consistency Detection}           & \multirow{2}{*}{Sum} \\ \cline{4-6}
			&                                   &                                   & Binarized Label Construction & Dynamic Point Identification & Total &                      \\ \hline
			$kitti\_1$          & 7.49                              & 27.98                             & 1.85           & 4.28                        & 6.13  & 41.60                \\
			$kitti\_2$          & 2.01                              & 24.57                             & 1.84           & 6.51                        & 8.35  & 34.93                \\
			$kitti\_3$          & 7.11                              & 36.07                             & 1.88           & 0.94                        & 2.82  & 46.00                \\
			$kitti\_4$          & 7.66                              & 18.93                             & 1.85           & 6.43                        & 8.28  & 34.87                \\
			$kitti\_5$          & 7.94                              & 19.67                             & 1.80           & 1.21                        & 3.01  & 27.61                \\
			$ulhk\_1$           & 5.45                              & 13.48                             & 0.53           & 3.87                        & 4.40  & 23.33                \\
			$ulhk\_2$           & 8.30                              & 13.20                             & 0.68           & 0.45                        & 1.13  & 21.50                \\
			$urban\_1$          & 6.07                              & 7.31                              & 1.27           & 1.81                        & 3.08  & 16.46                \\
			$urban\_2$          & 7.14                              & 7.03                              & 1.33           & 0.91                        & 2.24  & 16.41                \\
			$nclt\_1$           & 5.76                              & 21.78                             & 0.70           & 0.65                        & 1.35  & 28.89                \\
			$nclt\_2$           & 6.38                              & 23.12                             & 0.75           & 0.50                        & 1.25  & 30.75                \\
			$nclt\_3$           & 5.91                              & 22.20                             & 0.72           & 0.75                        & 1.47  & 29.58                \\
			$nclt\_4$           & 5.09                              & 20.31                             & 0.67           & 0.59                        & 1.26  & 26.66                \\
			$nclt\_5$           & 5.12                              & 19.81                             & 0.68           & 0.60                        & 1.28  & 26.21                \\
			$nclt\_6$           & 6.52                              & 20.48                             & 0.78           & 0.76                        & 1.54  & 28.54                \\
			$nclt\_7$           & 5.97                              & 24.43                             & 0.74           & 0.58                        & 1.32  & 31.72                \\
			$nclt\_8$           & 5.07                              & 19.64                             & 0.65           & 0.74                        & 1.39  & 26.10                \\
			$utbm\_1$           & 5.03                              & 14.23                             & 1.01           & 1.51                        & 2.52  & 21.78                \\
			$utbm\_2$           & 4.79                              & 16.91                             & 0.97           & 1.83                        & 2.80  & 24.50                \\
			$utbm\_3$           & 4.45                              & 14.27                             & 0.90           & 2.28                        & 3.18  & 21.90                \\
			$utbm\_4$           & 4.92                              & 14.51                             & 0.99           & 1.20                        & 2.19  & 21.62                \\
			$utbm\_5$           & 5.01                              & 15.14                             & 0.97           & 1.42                        & 2.39  & 22.54                \\
			$ulhk\_3$           & 6.54                              & 11.05                             & 0.98           & 0.39                        & 1.37  & 18.96                \\
			$ulhk\_4$           & 6.91                              & 9.22                              & 1.28           & 2.03                        & 3.31  & 19.44                \\ \hline
		\end{tabular}
	\end{center}
\end{table*}

\begin{table}[]
	\begin{center}
		\caption{Impact of Nearest Neighbor Search Method on Time Consumption of Label Consistency Detection (unit: ms)}
		\label{table33}
		\begin{threeparttable}
			\begin{tabular}{c|cc}
				\hline
				& \begin{tabular}[c]{@{}c@{}} LCD with 8-Nearest \\ Neighbor Search\end{tabular} & \begin{tabular}[c]{@{}c@{}}LCD with our \\ Nearest Neighbor Search\end{tabular} \\ \hline
				$kitti\_1$ & 78.53                                                                                                & \textbf{6.13}                                                                                                                    \\
				$kitti\_2$ & 115.99                                                                                               & \textbf{8.35}                                                                                                                    \\
				$kitti\_3$ & 81.04                                                                                                & \textbf{2.82}                                                                                                                    \\
				$kitti\_4$ & 43.22                                                                                                & \textbf{8.28}                                                                                                                    \\
				$kitti\_5$ & 46.63                                                                                                & \textbf{3.01}                                                                                                                    \\
				$ulhk\_1$  & 21.69                                                                                                & \textbf{4.40}                                                                                                                    \\
				$ulhk\_2$  & 21.58                                                                                                & \textbf{1.13}                                                                                                                    \\
				$urban\_1$ & 15.20                                                                                                & \textbf{3.08}                                                                                                                    \\
				$urban\_2$ & 15.50                                                                                                & \textbf{2.24}                                                                                                                    \\
				$nclt\_1$  & 8.76                                                                                                 & \textbf{0.65}                                                                                                                    \\
				$nclt\_2$  & 9.04                                                                                                 & \textbf{0.50}                                                                                                                    \\
				$nclt\_3$  & 8.78                                                                                                 & \textbf{0.75}                                                                                                                    \\
				$nclt\_4$  & 8.88                                                                                                 & \textbf{0.59}                                                                                                                    \\
				$nclt\_5$  & 9.35                                                                                                 & \textbf{0.60}                                                                                                                    \\
				$nclt\_6$  & 9.42                                                                                                 & \textbf{0.76}                                                                                                                    \\
				$nclt\_7$  & 9.05                                                                                                 & \textbf{0.58}                                                                                                                    \\
				$nclt\_8$  & 8.63                                                                                                 & \textbf{0.74}                                                                                                                    \\
				$utbm\_1$  & 11.50                                                                                                & \textbf{1.51}                                                                                                                    \\
				$utbm\_2$  & 10.56                                                                                                & \textbf{1.83}                                                                                                                    \\
				$utbm\_3$  & 11.44                                                                                                & \textbf{2.28}                                                                                                                    \\
				$utbm\_4$  & 12.47                                                                                                & \textbf{1.20}                                                                                                                    \\
				$utbm\_5$  & 10.92                                                                                                & \textbf{1.42}                                                                                                                    \\
				$ulhk\_3$  & 9.54                                                                                                 & \textbf{0.39}                                                                                                                    \\
				$ulhk\_4$  & 18.16                                                                                                & \textbf{2.03}                                                                                                                    \\ \hline
			\end{tabular}
		\end{threeparttable}
		\begin{tablenotes}
			\footnotesize
			\item[] \textbf{Denotations}: "LCD" is the abbreviation of "Label Consistency Detection".
		\end{tablenotes}
	\end{center}
\end{table}

Results in Table \ref{table32} demonstrate that our Dynamic-LIO outperforms state-of-the-arts for more than half sequences in terms of smaller ATE. Although IG-LIO achieves comparable results to our system on $nclt$ dataset, it shows poor accuracy on $utbm$ dataset. In addition, although our accuracy is not the best on $nclt\_2$, $nclt\_3$, $nclt\_5$, $nclt\_6$ and $ulhk\_4$, we are very close to the best accuracy. “-” means the corresponding value is not available. LIO-SAM needs 9-axis IMU data as input, while the $utbm$ dataset only provides 6-axis IMU data. Therefore, we cannot provide the results of LIO-SAM on the $utbm$ dataset. “$\times$” means the system fails to run entirely on the corresponding sequence. Except for our system, Fast-LIO2 and Point-LIO, other systems break down on several sequences, which also demonstrate the robustness of our system.

\subsection{Ablation Study of Undetermined-Points}
\label{Ablation Study of Undetermined-Points}

In our system, the purpose of incorporating undetermined-points is to remove dynamic points as much as possible, thereby increasing the proportion of static points in the output map. In this section, we validate the necessity of incorporating undetermined-points through comparing the PR and RR value of our Dynamic-LIO with and without considering undetermined-points.

Results in Table \ref{table4} and Table \ref{table5} demonstrate that incorporating undetermined-points can slightly improve PR and RR of our dynamic point detection and removal method.

\subsection{Ablation Study of Dynamic Point Removal for Pose Estimation}
\label{Ablation Study of Dynamic Point Removal for Pose Estimation}

In this section, we evaluate the effectiveness of removing dynamic points for pose estimation by comparing the ATE results of our Dynamic-LIO with and without removing dynamic points.

Results in Table \ref{table6} demonstrate that removing dynamic points can enhance the pose estimation accuracy of our Dynamic-LIO in dynamic scenes, especially on $urban-Nav$ dataset. In static scenes, our label consistency based dynamic point detection and removal has no negative effect on the pose estimation accuracy, which also reflects the robustness and practicability of the proposed method.

\subsection{Time Consumption Comparison with the State-of-the-Arts}
\label{Time Consumption Comparison with the State-of-the-Arts}

We compare the time consumption of our label consistency based dynamic point detection and removal method with two state-of-the-art online 3D point-based dynamic point detection and removal methods, i.e., Dynamic Filter \cite{fan2022dynamicfilter} and RH-Map \cite{yan2023rh}, on $semantic$-$kitti$ dataset \cite{behley2019semantickitti}. Then, we compare the time consumption of our Dynamic-LIO with RF-LIO and ID-LIO. Both results of other approaches are recorded from their literatures because they have not released the code. 

Results in Table \ref{table7} demonstrate that the time consumption of our dynamic point detection and removal method is much smaller than Dynamic Filter and RH-Map. The front-end of Dynamic Filter requires 55.71ms to process the data of a single sweep. When accounting for the back-end overhead, the total duration for processing a single sweep will be even longer. Given that the current LiDAR acquisition frequency is typically between 10$\sim$20Hz, this implies that the processing time for a single sweep must be within 50ms to ensure the real-time performance. It can be seen that neither Dynamic Filter nor RH-Map can guarantee real-time capability, whereas our method can rum in real time stably. Since the $semantic$-$kitti$ dataset does not include IMU data, we complete pose estimation and mapping in a LiDAR-only odometry mode while simultaneously detecting and removing dynamic objects online. Therefore, the time consumption recorded in Table \ref{table7} represents the total time cost for both LiDAR-only odometry and dynamic point detection and removal, and the results for Dynamic Filter and RH-Map are the same. Although the testing platforms are not entirely the same, the CPUs used for testing Dynamic Filter and RH-Map have a higher clock speed than the one we used. This also serves to a certain extent as evidence of the reference value of our experimental results. Results in Table \ref{table8} demonstrate that the time consumption of our Dynamic-LIO is much smaller than RF-LIO \cite{qian2022rf} and ID-LIO \cite{wu2023lidar}, while our system operates at a speed approximately 5X faster than that of RF-LIO and ID-LIO. Since RF-LIO is neither open-sourced nor tested on the $urban$-$Nav$ dataset, we are unable to obtain its results on sequence $urban\_1$ and $urban\_2$. As stated in ID-LIO \cite{wu2023lidar}, the authors tested their system using an i5 CPU. However, no specific model was provided in \cite{wu2023lidar}, thus we record the range of clock speeds for the entire i5 series of CPUs in Table \ref{table8}. Furthermore, although the i7-11700 has an advantage over the i5 series and i7-8559U in terms of thread count, our entire Dynamic-LIO system is implemented based on a single thread and does not take advantage of the multi-threading capability.

\begin{figure}
	\begin{center}
		\includegraphics[scale=0.605]{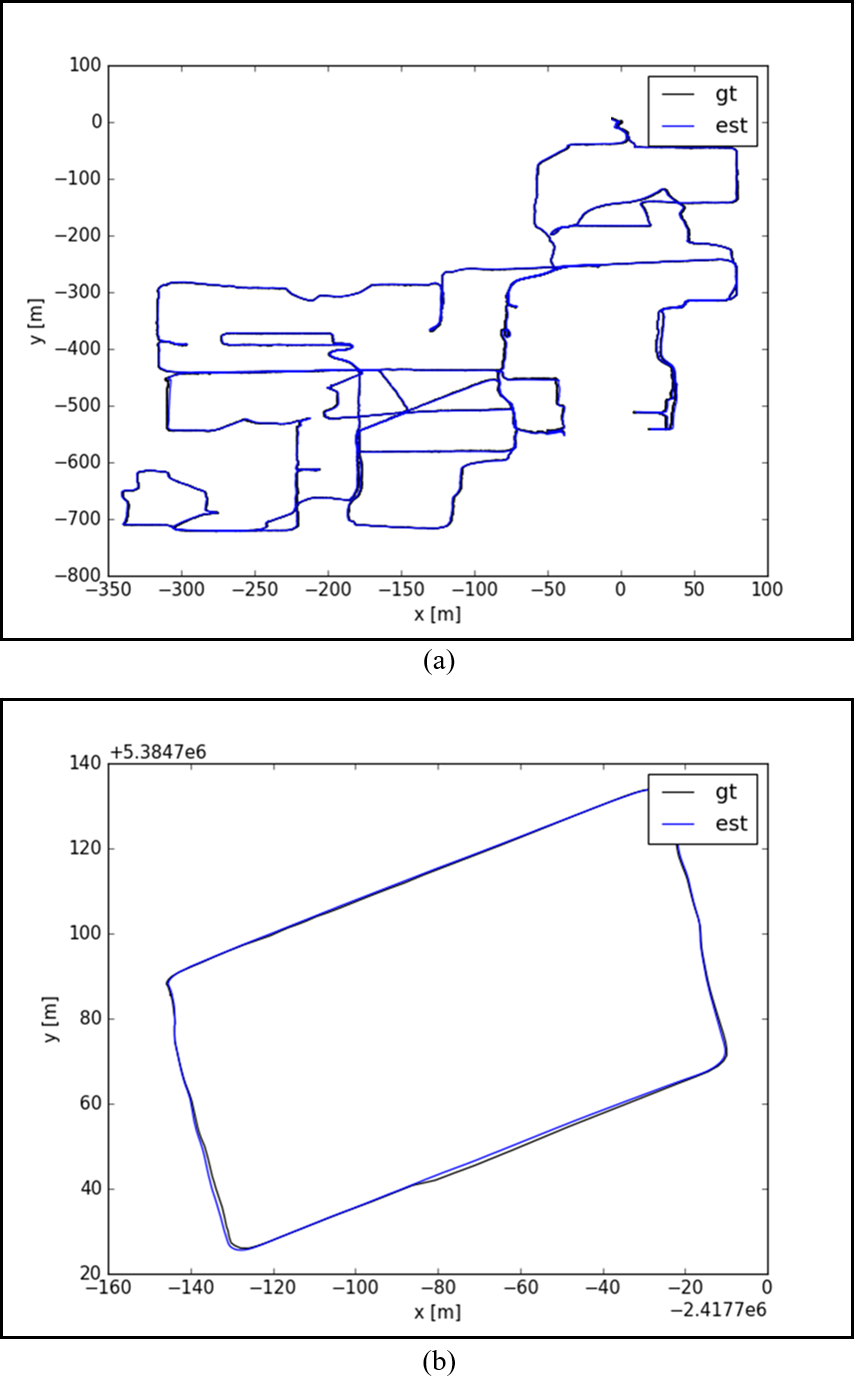}
		\caption{(a) and (b) are the comparison results between our estimated trajectories and ground truth on the exemplar sequences $nclt\_1$ and $ulhk\_3$.}
		\label{fig9}
	\end{center}
\end{figure}

\begin{figure}
	\begin{center}
		\includegraphics[scale=0.605]{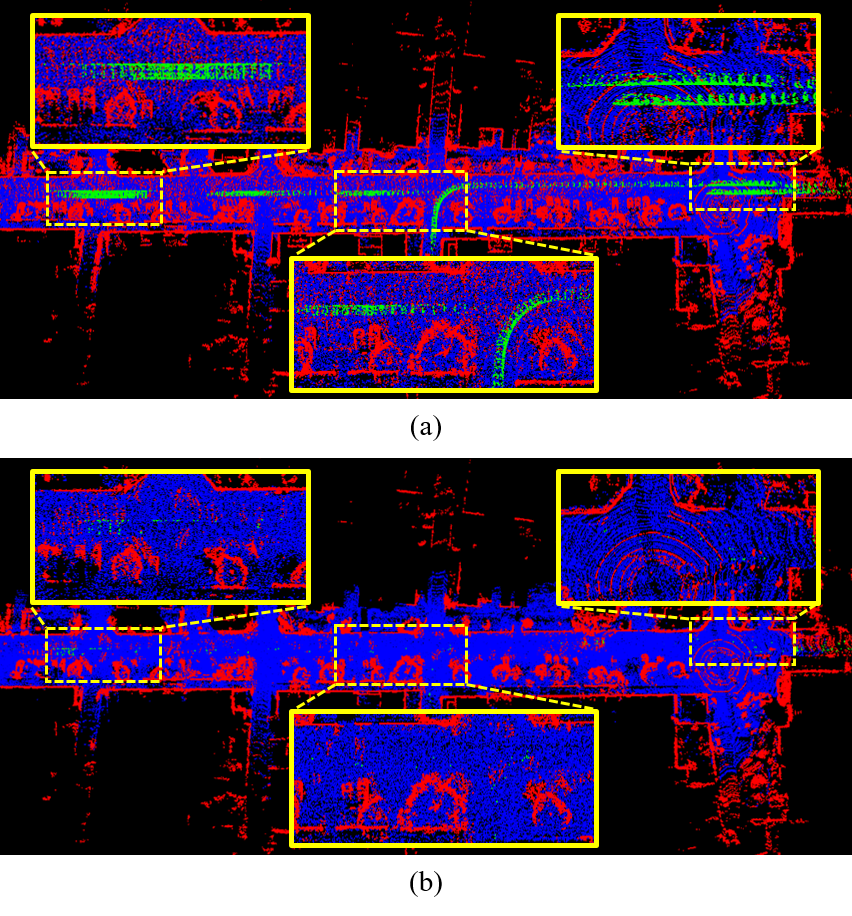}
		\caption{Visualization of (a) output map before removing dynamic points and (b) output map after removing dynamic points of the examplar sequence $kitti\_04$. The green points are the ghost tracks of moving objects.}
		\label{fig10}
	\end{center}
\end{figure}

\subsection{Time Consumption of Each Module}
\label{Time Consumption of Each Module}

We evaluate the runtime breakdown (unit: ms) of our system for all testing sequences. For each sequence, we test the time consumption of cloud processing (except for binarized label construction), state estimation and label consistency detection. The label consistency detection module can be further decomposed into two sub-steps: binarized label construction and dynamic point identification. Results in Table \ref{table9} show that our system takes only 1$\sim$9ms to identify dynamic points of a sweep, while the total duration for completing all tasks of LIO is 16$\sim$46ms. This implies that our method can accomplish the dynamic point detection and removal with extremely low computational overhead in LIO systems.

\subsection{Ablation Study of Nearest Neighbor Search}
\label{Ablation Study of Nearest Neighbor Search}

As mentioned in Sec. \ref{Voxel-Location-Based Nearest Neighbor Search}, compared to the conventional 8-nearest neighbor search, the voxel-location-based nearest neighbor search can significantly reduce the computational cost required for nearest neighbor search in label consistency detection. This subsection will provides quantitative comparative results to substantiate this conclusion.

Table \ref{table33} demonstrates that the voxel-location-based nearest neighbor search we employ has achieved an order-of-magnitude reduction in time consumption compared to the conventional 8-nearest neighbor search, primarily for the following two reasons: (1) The voxel-location-based nearest neighbor search only requires to process the voxel to which the current point belongs, whereas the 8-nearest neighbor search requires to process the voxel to which the current point belongs and its eight adjacent voxels; (2) The voxel-location-based nearest neighbor search does not necessitate the computation of the Euclidean distance between candidate points and the current point.

\subsection{Visualization for Trajectory and Map}
\label{Visualization for Map}

We visualize the trajectories and the local point cloud map estimated by our system. The comparison results between our estimated trajectories and ground truth of the exemplar sequences $nclt\_1$ and $ulhk\_3$ are shown in Fig. \ref{fig9} (a) and (b), where our estimated trajectories and ground truth almost exactly coincide. Fig. \ref{fig10} shows the ability of our Dynamic-LIO to reconstruct a static point cloud map on the exemplar sequence $kitti\_04$. As illustrated in Fig. \ref{fig10} (a), before removing dynamic points, the ghost tracks of moving objects (green points) are clearly visible on the map. As illustrated in Fig. \ref{fig10} (b), after removing dynamic points, the output map almost no longer contains ghost tracks.

\section{Conclusion}
\label{Conclusion}

This paper proposes a LIO system with label consistency detection, which can fastly eliminate the influence of moving objects in driving scenarios. Different from existing approaches involving batch geometric computation or global statistics to identify moving objects, the proposed method is more lightweight. Specifically, the proposed method constructs binarized labels for each point of current sweep, and utilizes the label difference between each point and its surrounding points in map to identify moving objects. Firstly, a fast 2D connected component method is employed to construct binarized labels, i.e., ground and non-ground, for each point of current sweep. Given that moving objects in driving scenes are located on ground, the inconsistency arises when comparing the label of a non-ground point in current sweep with its nearest neighbors in map. Meanwhile, we propose to utilize the voxel-location-based nearest neighbor to achieve fast nearest neighbor search. Furthermore, we embed the proposed label consistency detection method into a self-developed LIO, which can accurately estimate state and exclude the interference of dynamic objects with extremely low time consumption.

Experimental results show that the proposed label consistency detection method can achieve comparable PR and RR to state-of-the-art dynamic point detection and removal methods, while ensuring a lower computational cost. In addition, our Dynamic-LIO operates at a speed approximately 5X faster than state-of-the-art LIO systems for dynamic scenes, and achieve state-of-the-art pose estimation accuracy on both dynamic and static scenes.

% \section*{Acknowledgments}
% This should be a simple paragraph before the References to thank those individuals and institutions who have supported your work on this article.

\bibliographystyle{IEEEtrans}
\bibliography{IEEEabrv,IEEEExample}

\newpage

\vspace{11pt}

\begin{IEEEbiography}[{\includegraphics[width=1in,height=1.25in,clip,keepaspectratio]{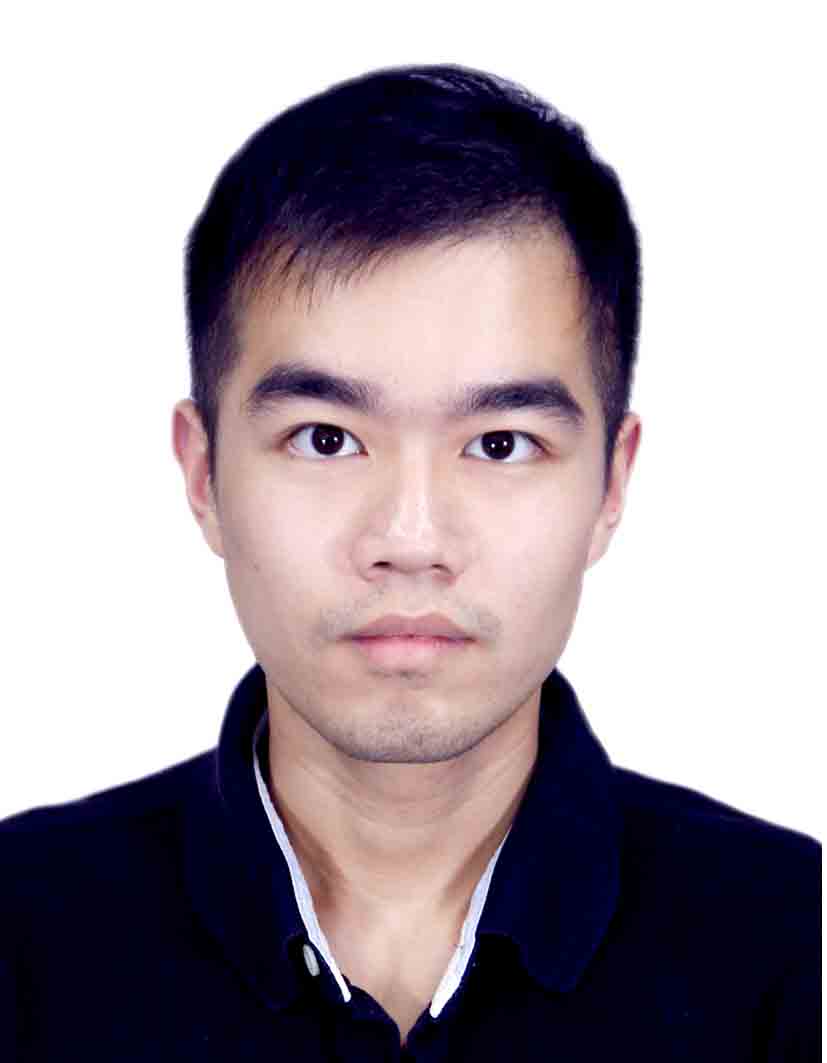}}]{Zikang~Yuan}
	received his PhD degree from Huazhong University of Science and Technology (HUST), Wuhan, China, in 2024. He has published one paper on TPAMI, one paper on RA-L, two papers on IROS, two papers on ACM MM and three papers on TMM. His research interests include monocular dense mapping, RGB-D simultaneous localization and mapping, visual-inertial state estimation, visual-LiDAR pose estimation and LiDAR-inertial state estimation.
\end{IEEEbiography}

\vspace{11pt}

\begin{IEEEbiography}[{\includegraphics[width=1in,height=1.25in,clip,keepaspectratio]{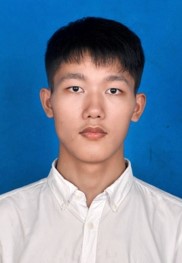}}]{Xiaoxiang~Wang}
	is currently a 2nd year M.S. student of Huazhong University of Science and Technology (HUST), School of Electronic Information and Communications. His research interests include 3D point based dynamic point indentification and object tracking.
\end{IEEEbiography}

\vspace{11pt}

\begin{IEEEbiography}[{\includegraphics[width=1in,height=1.25in,clip,keepaspectratio]{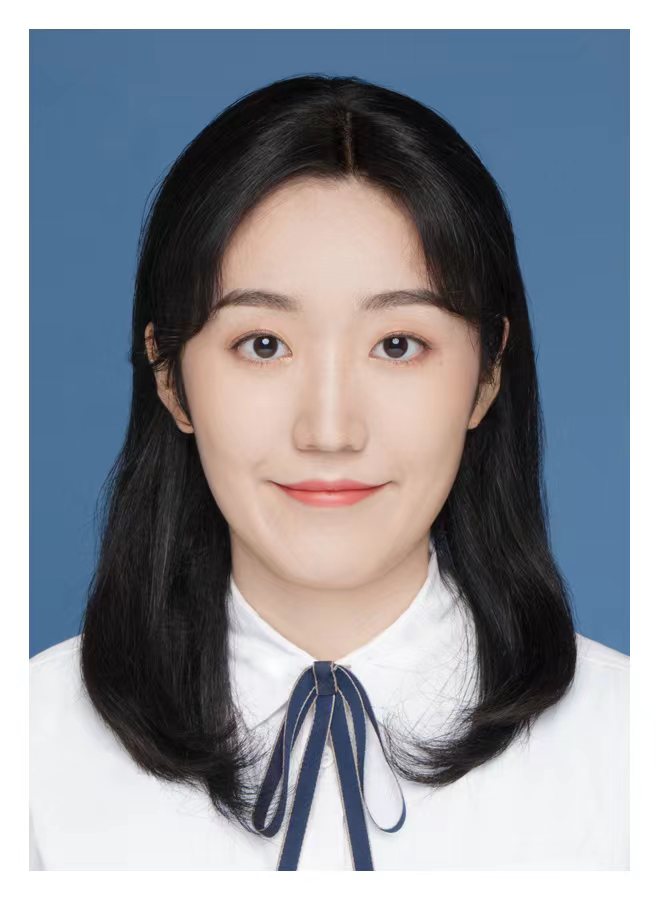}}]{Jingying~Wu}
	is currently a 3rd year M.S. student of Huazhong University of Science and Technology (HUST), School of Electronic Information and Communications. Her research interests include 3D point based dynamic point indentification and loop closure identification.
\end{IEEEbiography}

\vspace{11pt}

\begin{IEEEbiography}[{\includegraphics[width=1in,height=1.25in,clip,keepaspectratio]{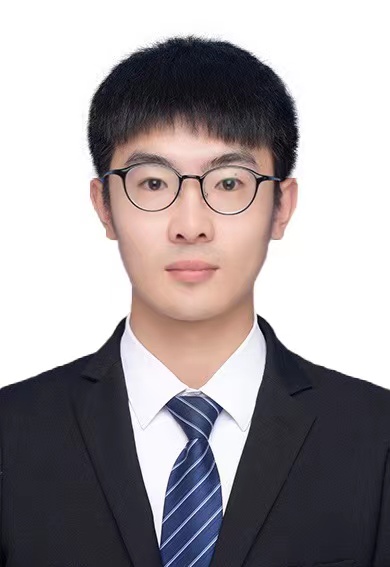}}]{Junda~Cheng}
	is currently a 2nd year PhD student at the Department of Electronic Information and Communications at Huazhong University of Science and Technology (HUST). He is supervised by Prof. Xin Yang. He received the B.Eng. degree from Huazhong University of Science and Technology in 2020. His research interests include stereo matching, multi-view stereo, and deep visual odometry.
\end{IEEEbiography}

\vspace{11pt}

\begin{IEEEbiography}[{\includegraphics[width=1in,height=1.25in,clip,keepaspectratio]{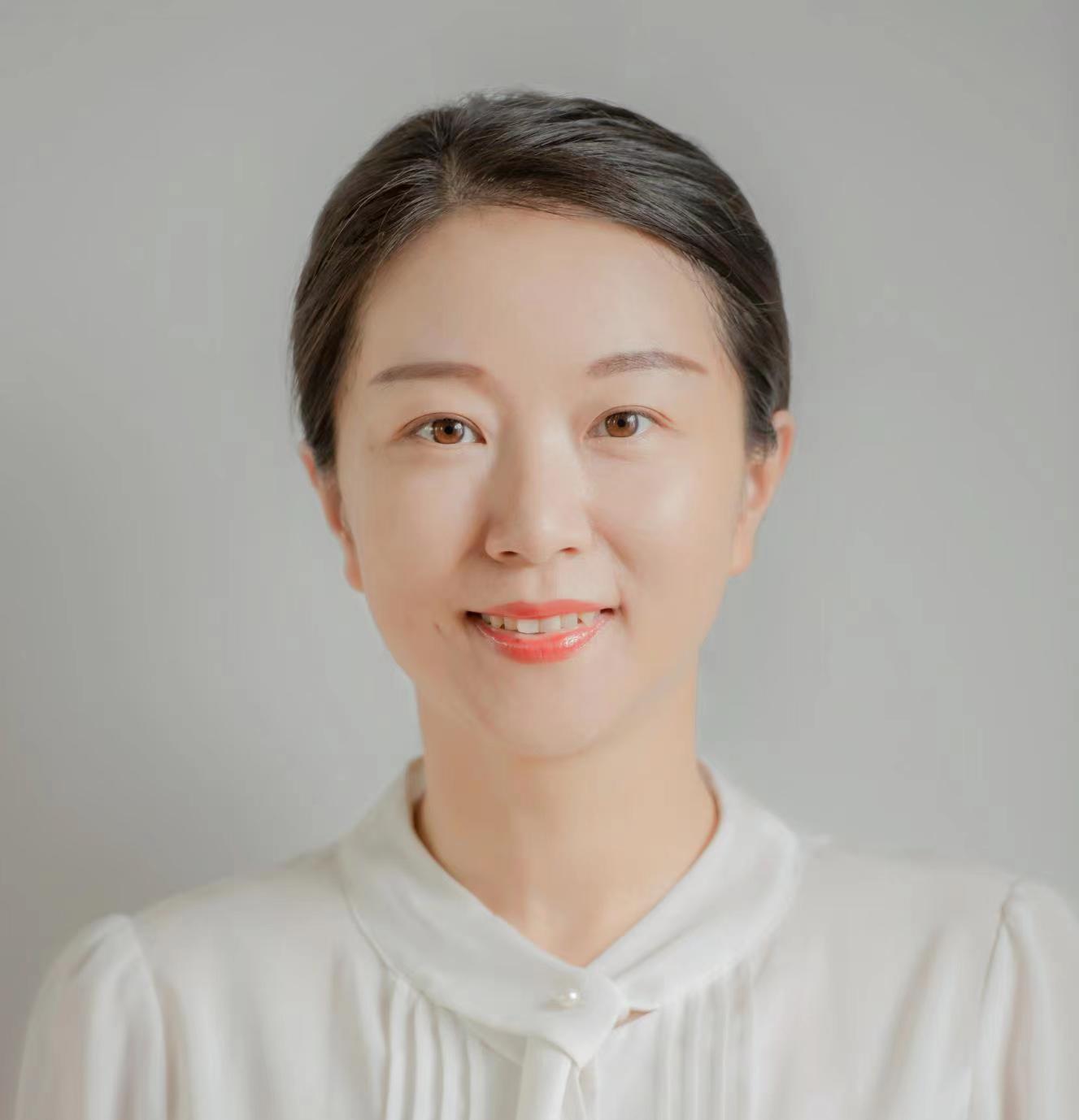}}]{Xin~Yang}
	received her PhD degree in University
	of California, Santa Barbara in 2013. She worked
	as a Post-doc in Learning-based Multimedia Lab at
	UCSB (2013-2014). She is current Professor
	of Huazhong University of Science and Technology
	School of Electronic Information and Communications.
	Her research interests include
	simultaneous localization and mapping, augmented
	reality, and medical image analysis. She has published
	over 90 technical papers, including TPAMI, IJCV,
	TMI, MedIA, CVPR, ECCV, MM, etc., co-authored two books and holds 3 U.S. Patents. Prof.
	Yang is a member of IEEE and a member of ACM.
\end{IEEEbiography}

\vfill

\end{document}